\documentclass[12pt,a4paper]{ijicic}

\def\currentvolume{7}
\def\currentissue{1}
\def\currentyear{2011}
\def\currentmonth{January}
\def\ppages{1--10}

\newtheorem{lemma}{Lemma}

\usepackage[dvips]{graphicx}
\usepackage{epsfig}
\usepackage{epsf}

\usepackage{furumath}
\usepackage{color}
\usepackage{amssymb}
\usepackage{url}

\newcommand{\du}{\,\mathrm{d}u}
\newcommand{\dv}{\,\mathrm{d}v}
\newcommand{\dw}{\,\mathrm{d}w}
\newcommand{\IP}{\mathrm{I}_P}
\newcommand{\Ip}{\mathrm{I}_p}
\newcommand{\Iq}{\mathrm{I}_q}

\title[Heterogeneous co-occurrence embedding]
      {Heterogeneous co-occurrence embedding for visual information exploration}
\author[T. Ishida and T. Furukawa]{}

\begin{document}
\maketitle

\centerline{\scshape  Takuro Ishida$^1$ and Tetsuo Furukawa$^1$}
 \medskip
{\footnotesize
\centerline{$^1$Graduate School of Life Science and Systems Engineering}
\centerline{Kyushu Institute of Technology} 
\centerline{2--4 Hibikino, Wakamatsu-ku,
Kitakyushu-shi, Fukuoka, Japan}
\centerline{ishida.takuro249@mail.kyutech.jp; furukawa@brain.kyutech.ac.jp} }
\medskip
\medskip


\begin{abstract}
{\em This paper proposes an embedding method for co-occurrence data aimed at visual information exploration. We consider cases where co-occurrence probabilities are measured between pairs of elements from heterogeneous domains. The proposed method maps these heterogeneous elements into corresponding two-dimensional latent spaces, enabling visualization of asymmetric relationships between the domains. The key idea is to embed the elements in a way that maximizes their mutual information, thereby preserving the original dependency structure as much as possible. This approach can be naturally extended to cases involving three or more domains, using a generalization of mutual information known as total correlation. For inter-domain analysis, we also propose a visualization method that assigns colors to the latent spaces based on conditional probabilities, allowing users to explore asymmetric relationships interactively. We demonstrate the utility of the method through applications to an adjective–noun dataset, the NeurIPS dataset, and a subject–verb–object dataset, showcasing both intra- and inter-domain analysis.}\\
{\bf Keywords:} co-occurrence data; embedding method; mutual information maximization; visual information retrieval.
\end{abstract}

\section{Introduction}
Co-occurrence data represent the frequency with which two elements are observed together in the same event. A typical example is word co-occurrence data, which measures how often pairs of words appear in the same sentence~\cite{Dagan199943,Weeds2005439}. Other examples include the co-occurrence of animal species in the same environment~\cite{Sanderson2009771}, co-citation and co-authorship in academic articles~\cite{Small1973265,Leydesdorff20061616,Otte2002441}, co-occurrence of songs in playlists~\cite{Chen2012714}, email traffic between users~\cite{Fu20071}, and co-occurrence of diseases~\cite{Wang2021}.

To represent the structure behind co-occurrence data, embedding approaches are widely used. These methods project elements into a low-dimensional latent space, where frequently co-occurring elements are placed close together. In this space, each element is represented as a vector, allowing for geometric and statistical interpretation. When the latent space is two-dimensional, such embeddings enable direct visualization of inter-element relationships, helping analysts intuitively grasp co-occurrence patterns and uncover latent associations. Common techniques include multi-dimensional scaling (MDS), which interprets co-occurrence frequencies as proximity measures~\cite{Leydesdorff20061616}, and graph-based methods such as Isomap, which leverage Laplacian eigenmaps~\cite{Mahecha200931}.

Co-occurrence relationships are not limited to elements within a single domain; they are often defined between elements from different domains as well. Typical examples include adjective–noun pairs in text~\cite{Kohara201347} and image–hashtag pairs on websites~\cite{Zhang2016,Liu2015867}.
In document analysis, datasets represented by conditional probabilities such as the bag-of-words (BoW) representation can also be interpreted as a form of heterogeneous co-occurrence~\cite{Globerson20072265}.
Moreover, some co-occurrence relationships involve asymmetric roles between elements within the same domain. For example, questioner–questionee pairs in QA systems~\cite{Fang2016122}, citing–cited relations in academic publications~\cite{Leydesdorff20061616}, and follower–followee relationships in social media~\cite{Liu20161774} reflect directionality or role differentiation. These are often treated as heterogeneous in practice due to their inherent asymmetry.
Co-occurrence can also involve more than two domains, such as subject–verb–object triplets in sentence structures~\cite{Kohara201347}.

For heterogeneous co-occurrence data, embedding approaches typically aim to map all elements from different domains into a single shared latent space, a process often referred to as \emph{co-embedding}.  
A well-known example is the Co-occurrence Data Embedding (CODE) method~\cite{Globerson20072265}, which assumes that the logarithm of the co-occurrence probability is proportional to the Euclidean distance between elements in the latent space. CODE has been applied in various domains, and several extensions have been proposed~\cite{Sarkar2007420,Maron2010nips,Khoshneshin201087,Khoshneshin201174}.  
Co-embedding enables intuitive visualization by placing all elements in the same space. However, this convenience comes at a cost: it tends to obscure the asymmetric relationships between domains, which are often crucial for understanding the structure of the data.  

To address this limitation, this study maps elements to separate latent spaces for each domain, rather than embedding them into a shared space.  
While this approach allows for the representation of asymmetric relationships across heterogeneous domains, it introduces two technical challenges.  
First, since elements are embedded in distinct spaces, the distance between elements from different domains is no longer defined, requiring an alternative means of determining their latent variables.  
Second, the use of multiple spaces calls for new visualization methods to help analysts interpret inter-domain relationships.  
To tackle these issues, we propose a method that estimates latent variables and co-occurrence density functions such that the mutual information between domains is maximally preserved in the embedding space.  
In particular, we demonstrate that our objective function minimizes a variational upper bound on the KL divergence between the observed and modeled co-occurrence, thereby providing a theoretical guarantee for the optimality of the latent structure.

The main contributions of this study are as follows:
\begin{itemize}
  \item We propose a domain-specific embedding method that represents asymmetric co-occurrence relationships by mapping each domain into a separate latent space.
  \item The method is theoretically grounded, as it maximizes a variational upper bound corresponding to the KL divergence between observed and modeled co-occurrence.
  \item We provide a novel visualization framework that supports interactive and iterative exploration, in contrast to conventional methods that offer only static views of inter-element relationships.
\end{itemize}

The remainder of this paper is organized as follows. Section~2 reviews background and related work. Section~3 introduces the proposed method. Section~4 describes the visualization framework, and Section~5 presents experimental results. Section~6 discusses the findings, and Section~7 concludes the paper.

\section{Background and related work}

\subsection{Aims and roles of co-occurrence embedding}

The aim of co-occurrence embedding is to represent elements as vectors in a latent space, capturing their co-occurrence relationships. Broadly speaking, the objectives of co-occurrence embedding can be divided into two categories.
The first group focuses on obtaining vector representations for downstream applications such as classification, prediction, or recommendation. In this setting, the key requirement is that elements can be treated as feature vectors, making linear operations and machine learning tasks tractable. Embeddings are typically constructed in a middle-dimensional space, which is smaller than the number of elements but large enough to retain predictive capacity, often in the range of a hundred dimensions or more. This line of work is generally known as \emph{representation learning}~\cite{Xie2021}, and has been widely studied in recent deep embedding models.

The second group aims at visualization, interpretation, and exploratory analysis of the co-occurrence structure. Here, the goal is not to improve prediction accuracy, but rather to reveal interpretable relationships and patterns within the data. The latent space must therefore be low-dimensional---typically two or three dimensions---so that the resulting embeddings can be directly visualized. This perspective emphasizes the modeling of interpretable structures over numerical accuracy, and often employs techniques such as dimensionality reduction or manifold learning~\cite{Globerson20072265,Mahecha200931,Leydesdorff20061616}.

This study belongs to the second group. While recent deep embedding and attention-based methods belong to the first category, we emphasize that their objectives are fundamentally different from ours. As such, performance comparisons between the two are not meaningful, since they target different goals.

\subsection{Representative methods of co-occurrence embedding}
\label{sec:related_representative}

Existing approaches for embedding co-occurrence data can be classified by the modeling principles and their target dimensionality: 
(i) \emph{classical spectral methods}, 
(ii) \emph{shallow matrix/tensor factorization}, and 
(iii) \emph{neural approach and deep embeddings}, 
which respectively prioritize visualization, mid-dimensional feature learning, and predictive performance. We briefly review each family below.

\subsubsection*{\bf Spectral methods.}
Early work, such as classical multi-dimensional scaling (MDS)~\cite{White1981163} and its non-linear extensions (e.g.\ Isomap, Laplacian Eigenmaps, Diffusion Maps), treats a co-occurrence matrix as a proximity matrix and seeks an embedding that satisfies $P_{ij}\propto\exp\bigl(-\|u_i-u_j\|^2\bigr)$.  These methods excel at revealing global geometry in low (2--3) dimensions, but they assume pairwise proximity and are hard to extend to higher-order or heterogeneous co-occurrence.

\subsubsection*{\bf Matrix and tensor factorization.}
A widely used strategy first converts raw frequencies to \emph{shifted positive pointwise mutual information} (SPPMI)~\cite{Levy20142177}, $\tilde M_{ij} = \max\bigl(\log\tfrac{P_{ij}}{P_iP_j}-\log\alpha,0\bigr)$, and then applies truncated SVD, LSA, or non-negative matrix factorisation to obtain $\tilde \vM \approx \vU\vV^{\mathsf T}$, where $\vU$ and $\vV$ can differ when the co-occurrence matrix is asymmetric.  GloVe~\cite{Pennington20141532} popularised this pipeline for word embeddings, and extensions to higher-order tensors (CP/Tucker decompositions) handle 3-way or multi-domain interactions~\cite{VanDeCruys2010417,Nickel2011809}.  
Such methods yield compact (50--300 d) features effective for downstream tasks, but offer limited control over the interpretability of the geometry they learn.

\subsubsection*{\bf Neural approaches and deep embeddings.}
Neural approaches learn embeddings by directly modeling co-occurrence signals.  
The skip-gram model with negative sampling (SGNS)~\cite{Mikolov2013} interprets word--context pairs as positive co-occurrences and learns high-dimensional representations.  
This framework has been extended to graph-structured co-occurrence data through random-walk-based methods such as LINE~\cite{Tang20151067} and node2vec~\cite{Grover2016855}.  
More recently, inductive graph neural networks such as GraphSAGE~\cite{Hamilton20171025} and the Graph Attention Network (GAT)~\cite{Velickovic2018} have been proposed to aggregate, or attentively weight, local neighborhoods in a scalable manner.  

\subsubsection*{\bf Relationship to this study.}
Unlike earlier work that pursues either \emph{high-accuracy prediction} with deep embeddings or \emph{mid-dimensional feature construction} through matrix/tensor factorization, our goal is to \emph{capture and visualize asymmetric co-occurrence across heterogeneous domains in a genuinely low-dimensional latent map}. Concretely, we project entities into a two- to three-dimensional nonlinear manifold whose kernel-density formulation preserves point-wise mutual information while revealing domain-specific structure. Deep-learning and factorization methods therefore serve as complementary baselines rather than direct competitors.

\begin{figure}
  \centering
  \begin{minipage}[c]{0.48\linewidth}
    \centering
    \includegraphics[width=\linewidth]{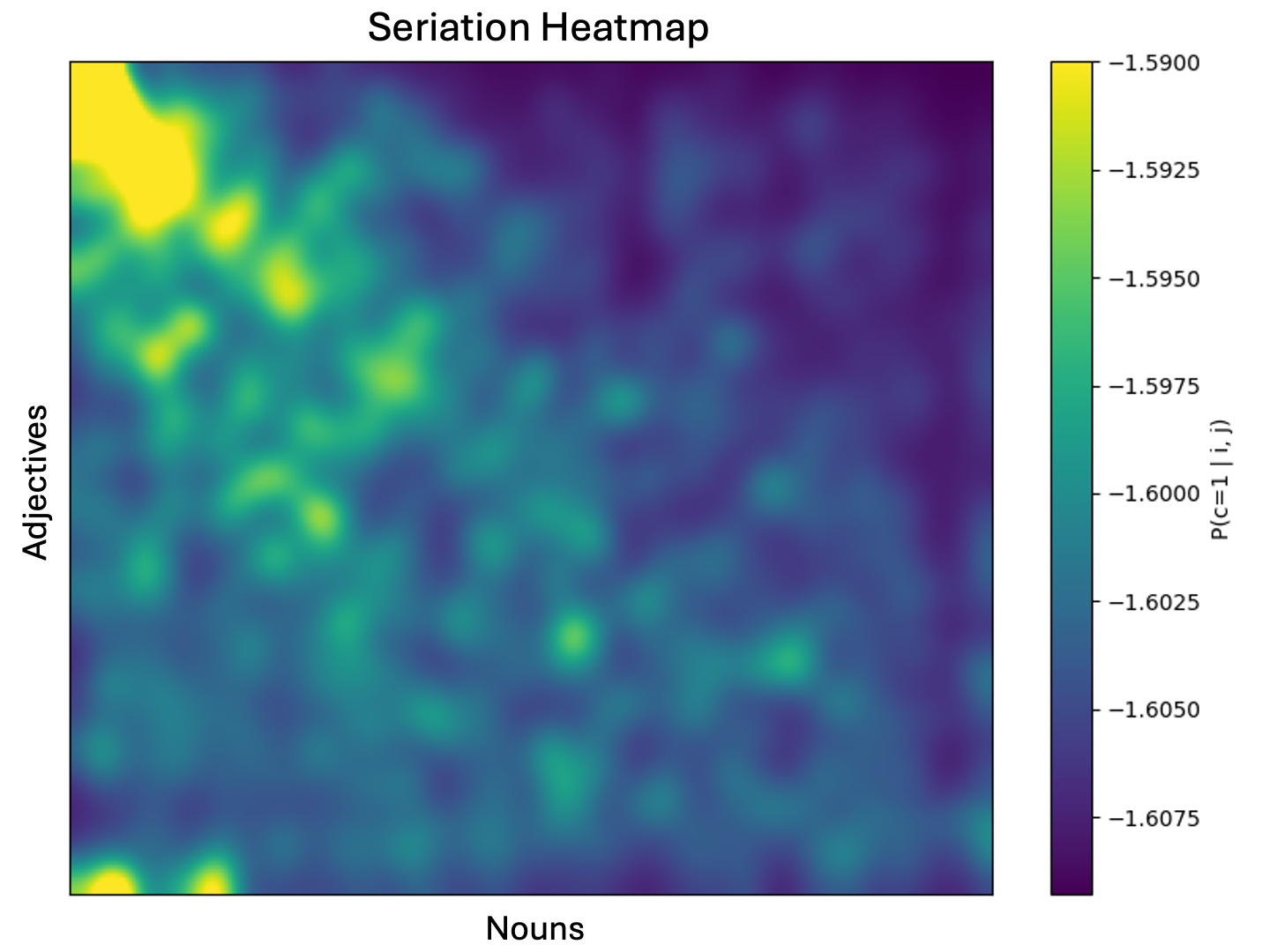}\\
    (a)
  \end{minipage}
  \begin{minipage}[c]{0.48\linewidth}
    \centering
    \includegraphics[width=\linewidth]{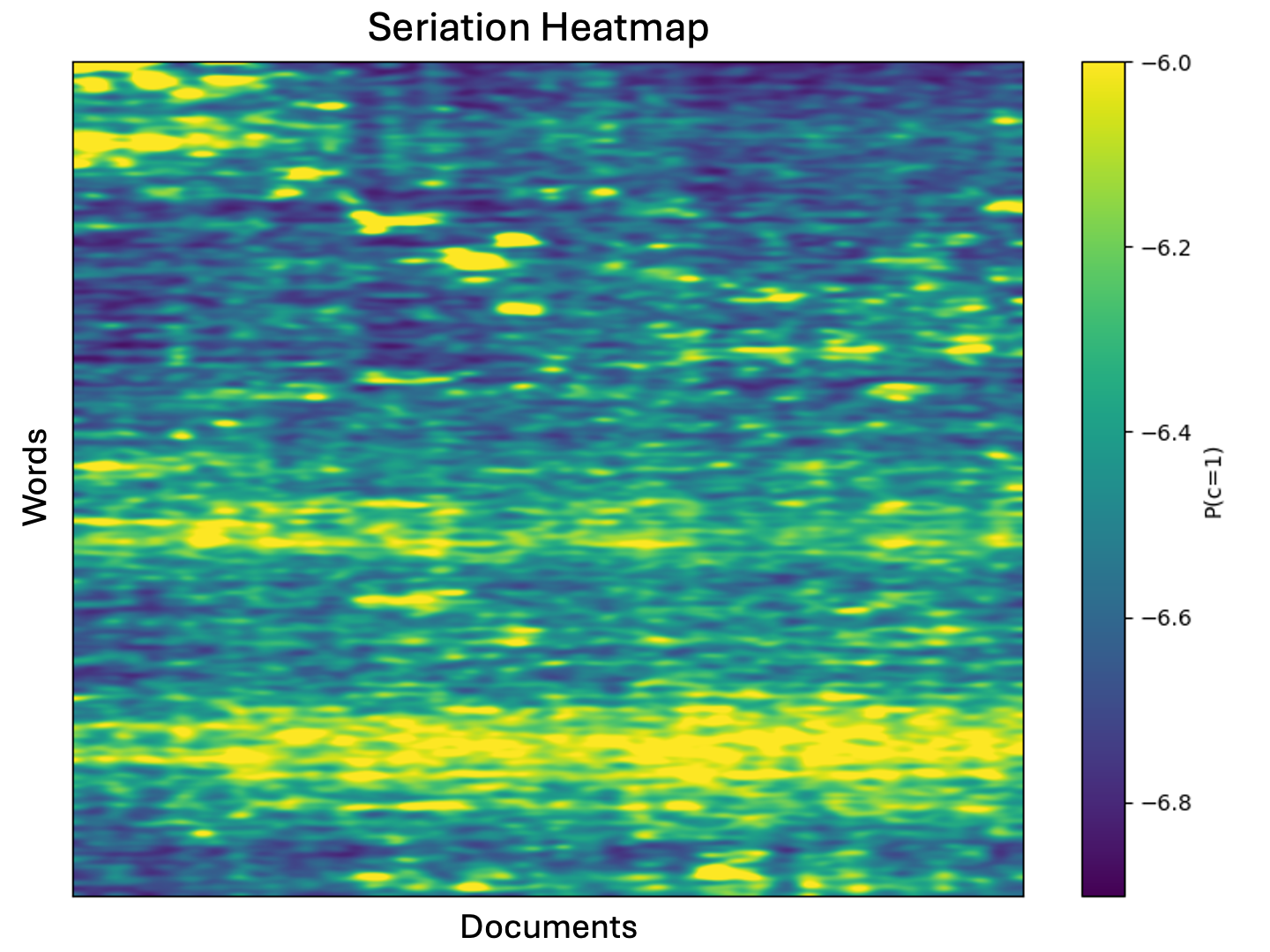}\\
    (b)
  \end{minipage}
  \caption{Seriation heatmaps of heterogeneous co-occurrence data. (a) Adjective–noun co-occurrence shows a clear diagonal structure, reflecting strong alignment between specific adjectives and nouns. Notably, off-diagonal spots in the lower-left region suggest asymmetric relationships where a noun co-occurs with multiple adjective clusters. (b) Document--term co-occurrence from NeurIPS papers exhibits more complex asymmetric patterns. Diagonal traces correspond to topic alignment, while off-diagonal structures and horizontal bands indicate cross-domain term usage and general-purpose vocabulary, respectively.}
\figlabel{seriation}
\end{figure}

\subsection{Asymmetric Structures in Heterogeneous Co-occurrence}
\label{sec:asymmetry}

Heterogeneous co-occurrence data, such as adjective–noun, paper–term, or follower–followee pairs, differ from single-domain co-occurrence data in one key aspect: \textbf{directionality and role asymmetry}. While single-domain co-occurrence is typically symmetric (for example, two words appearing in the same sentence), heterogeneous relations often involve directionality or role differences. Capturing this asymmetry is essential for modeling the true structure of such data \cite{Gries2013137}.

To illustrate this, we begin with adjective--noun co-occurrence data. In this setting, each adjective tends to modify a specific subset of nouns, resulting in a clear alignment between the two sets. When the co-occurrence matrix is rearranged using seriation, a dominant diagonal structure emerges, reflecting this alignment (\figref{seriation}~(a)).  
However, closer inspection reveals asymmetric patterns as well. For instance, in the lower-left corner of the seriation heatmap, we observe off-diagonal spots where a particular noun co-occurs with two distinct clusters of adjectives. This suggests that the noun participates in multiple semantic contexts, leading to asymmetric relationships across clusters, and such directional contexts are known to improve the modeling of functional similarity \cite{Levy20142177}. Identifying and modeling such off-diagonal connections is precisely the kind of structural asymmetry our approach aims to capture.

In more complex settings such as paper--term co-occurrence in scientific literature, asymmetric structures become more prominent. For example, papers in reinforcement learning often include terms related to online optimization, even though those terms are also commonly used in other contexts such as bandit algorithms. This leads to off-diagonal co-occurrence patterns that reflect asymmetric semantic relations between domains. \figref{seriation}~(b) shows an example based on NeurIPS paper--term data, where such asymmetric features are visibly pronounced. 

In addition to these localized asymmetries, some terms appear broadly across many domains and serve as general-purpose vocabulary. These words tend to occur across all papers regardless of their primary topics, forming horizontal bands in the heatmap. This phenomenon is analogous to ``influencers'' in social networks---nodes that are widely followed without strong reciprocity---playing a connective role across otherwise independent clusters.

These types of asymmetric structures have been widely discussed in linguistic and semantic modeling literature. For example, Levy \& Goldberg \cite{Levy2014302} introduced dependency-based embeddings, which leverage syntactic relations to capture asymmetric functional similarities between words---showing clear advantages over symmetric window-based models. Similarly, Roller et al.\ \cite{Roller20141025} and Tissier at al.\ \cite{Tissier2017254} argued that asymmetric co-occurrence signals (such as subject--object or modifier--head dependencies) provide critical information for modeling selectional preferences and predicate-argument structures.

These studies highlight the importance of capturing \textbf{asymmetry not only as a data artifact but as a semantic signal}. While previous work has focused primarily on linguistic corpora, our study extends this perspective to a broader class of heterogeneous co-occurrence domains, including scientific papers, social networks, and visual-semantic relationships.

Building on these insights, our method aims to make asymmetric structures both formal and visible. Specifically, we (1) embed each domain into a distinct latent space, (2) model the asymmetric co-occurrence probability across them, and (3) reveal interpretable visual patterns that emphasize off-diagonal and role-specific alignments.

\subsection{Existing approaches for heterogeneous co-occurrence embedding and their limitations}

Most existing methods for heterogeneous co-occurrence embedding are based on either distance-based models or matrix factorization. A well-known example of the former is CODE~\cite{Globerson20072265}, which assumes $P_{ij} \propto \exp(-|u_i - v_j|^2)$, where $u_i$, $v_j$ are embedding vectors. An example of the latter is SPPMI-based matrix decomposition~\cite{Liang201659}, which assumes $\text{SPPMI}_{ij} \approx u_i^\top v_j$.

Both approaches embed heterogeneous elements $i$ and $j$ into a shared vector space and use symmetric similarity functions. While this allows for efficient modeling, it inherently limits the ability to capture directional or role-specific relationships that often arise in heterogeneous co-occurrence data.

To overcome this limitation, we propose embedding heterogeneous elements into separate latent spaces. Specifically, for two distinct sets $A$ and $B$, each element $a_i \in A$ and $b_j \in B$ is mapped to a vector $u_i \in \mathcal{U}$ and $v_j \in \mathcal{V}$, where $\mathcal{U} \neq \mathcal{V}$. This separation allows us to explicitly represent asymmetry between domains. Furthermore, instead of relying on symmetric measures such as Euclidean distance or inner product, we directly model the co-occurrence probability density $q(u, v)$. This enables a more expressive representation of directional and role-dependent relationships.

\section{Proposed method}

\subsection{Theoretical framework: Problem formulation and lemmas}

Assume two sets $A=\{a_i\}$ and $B=\{b_j\}$, with the empirical co-occurrence probability of $(a_i,b_j)$ being $P(a_i,b_j)\equiv P_{ij}$. First, we embed $a_i\in A$ to $u_i\in\cU$, and embed $b_j\in B$ to $v_j\in\cV$, where $\cU=\bbR^{d_A}$ and $\cV=\bbR^{d_B}$ are the embedding spaces of set $A$ and $B$. We aim to estimate the probability densities $p(u\mid a_i)$ and $p(v\mid b_j)$ in the embedding spaces $\cU$, $\cV$, instead of assigning a single point estimate to each element in the latent space. This study assumes that the densities are represented as $p(u\mid a_i)=: k(u\mid u_i)=\cN(u\mid u_i,\sigma^2\vI)$, and $p(v\mid b_j)=:k(v\mid v_j)=\cN(v\mid v_j,\sigma^2\vI)$. The dimensions of two embedding spaces, $d_A$ and $d_B$, do not necessarily need to be equal in our framework. 

Second, we estimate the joint probability density $q(u,v)$, which represents the co-occurrence density of $(u,v)$ in the product embedding space $\cU\times\cV$. We define $q(u,v)$ as the modeled co-occurrence density. The modeled co-occurrence probability of $a_i$ and $b_j$ is represented as
\begin{align}
    Q_{ij}\equiv Q(a_i,b_j)&=\iint P(a_i\mid u)\,P(b_j\mid v)\,q(u,v)\du \dv.
    \eqlabel{Qij1}
\end{align}
$Q(a_i,b_j)$ depends not only on $U=\{u_i\}$ and $V=\{v_j\}$, but also on $q(u,v)$. This relationship contrasts with conventional methods, in which $Q(a_i,b_j)$ depends on only $u_i$ and $v_j$ (except the constant coefficient). 

In applying Bayes' theorem, \eqref{Qij1} is represented as
\begin{align}
    Q(a_i,b_j) &=\iint \frac{p(u\mid a_i)P(a_i)}{p(u)}\,
      \frac{p(v\mid b_j)P(b_j)}{p(v)}\,q(u,v)\,
      \du\dv \notag\\
          &=P(a_i)\,P(b_j)\iint k(u \mid u_i)\,k(v \mid v_j)\,
      \frac{q(u,v)}{p(u)p(v)}\,
      \du\dv,
   \eqlabel{Qij2}
\end{align}
where $p(u)=\sum_i k(u \mid u_i)P(a_i)$ and $p(v)=\sum_j k(v \mid v_j)P(b_j)$. Our aim is to estimate $U=\left\{u_i\right\}$, $V=\left\{v_j\right\}$ and $q(u,v)$, so that the modeled co-occurrence probability $Q(a_i,b_j)$ approximates the empirical co-occurrence probability $P(a_i,b_j)$. If the kernel width $\sigma$ is sufficiently small, the integral in \eqref{Qij2} can be approximated using the first-order delta method, yielding:
\begin{align*}
    Q(a_i,b_j)&\simeq P(a_i)\, P(b_j) \,\frac{q(u_i,v_j)}{p(u_i)\,p(v_j)}.
\end{align*}
Because our aim is to embed elements so that $P(a_i,b_j)\approx Q(a_i,b_j)$, our task is to estimate $U$, $V$ and $q(u,v)$ so that $P(a_i,b_j)$ is approximated as
\begin{align*}
    \frac{P(a_i,b_j)}{P(a_i)\,P(b_j)} \approx \frac{q(u_i,v_j)}{p(u_i)\,p(v_j)}.
\end{align*}
This implies that modeling co-occurrence probability amounts to estimating a ratio of densities in the embedding space.

The following lemmas exist under this problem formulation:
\begin{lemma}
  The upper bound of Kullback-Leibler (KL) divergence between $P(a_i,b_j)$ and $Q(a_i,b_j)$ is given by 
\begin{align}
    \DKL{P}{Q} \leq \IP[a;b] - \Ip[u;v] + \DKL{p}{q},
  \eqlabel{lemma1}
\end{align}
where
\begin{align}
    \IP[a;b] &= \sum_{i,j} P(a_i,b_j)\log\frac{P(a_i,b_j)}{P(a_i)P(b_j)}, \nonumber \\
    \Ip[u;v] &= \iint p(u,v)\log\frac{p(u,v)}{p(u)p(v)}\,du\,dv, \nonumber \\
    p(u,v) &= \sum_{i,j}k(u \mid u_i)\,k(v \mid v_j)P(a_i,b_j). \eqlabel{eq:kernel_density}
\end{align}
    
\end{lemma}
{\bf Proof: }By applying \eqref{Qij2} and Jensen's inequality, the upper bound is given by
\begin{align*}
     \DKL{P}{Q} =& -\sum_{i,j} P(a_i,b_j) \log\frac{Q(a_i,b_j)}{P(a_i,b_j)} \\
    =& -\sum_{i,j} P(a_i,b_j) \log 
       \left(
         \frac{P(a_i)\,P(b_j)}{P(a_i,b_j)}
         \iint k(u \mid u_i)\,k(v \mid v_j)\frac{q(u,v)}{p(u)p(v)}
         \du \dv
       \right) \\
    =& \sum_{i,j} P(a_i,b_j)\log\frac{P(a_i,b_j)}{P(a_i)P(b_j)} \\
      &-\sum_{i,j}P(a_i,b_j)\log
        \left(
          \iint k(u \mid u_i)\,k(v \mid v_j) \frac{q(u,v)}{p(u)p(v)}
          \du \dv
        \right) \\
    \leq& \IP[a;b]-\sum_{i,j}\iint k(u \mid u_i)\,k(v \mid v_j)P(a_i,b_j)
      \log \frac{q(u,v)}{p(u)p(v)}\du \dv \\
    =& \IP[a;b] -\iint p(u,v)\log 
      \frac{q(u,v)}{p(u,v)}\frac{p(u,v)}{p(u)p(v)} \du\dv\\
    =& \IP[a;b]-\Ip[u;v] +\DKL{p}{q}. 
\end{align*} 
(See Appendix A for a full step-by-step derivation.)\\
\qed

\begin{lemma}
  When $U=\{u_i\}$ and $V=\{v_j\}$ are given, the upper bound of $\DKL{P}{Q}$ is minimized with respect to $q$ if and only if $q$ satisfies
  \begin{align}
      q(u,v)=p(u,v)=\sum_{i,j} k(u\mid u_i)\,k(v\mid v_j)\,P(a_i,b_j). 
    \eqlabel{lemma2}
  \end{align}
  If $q$ satisfies this condition, the upper bound of $\DKL{P}{Q}$ is given as $\DKL{P}{Q}\leq\IP[a;b]-\Iq[u;v]$.
\end{lemma}
{\bf Proof: }
  $\DKL{p}{q}=0$ in \eqref{lemma1} if and only if $p=q$, and $\Ip[u;v]=\Iq[u;v]$ under this condition.\\ \qed \\

Thus, when the embedding $U=\{u_i\}$ and $V=\{v_j\}$ are given, the modeled co-occurrence density $q(u,v)$ is uniquely determined by the kernel density estimation---by regarding $k(u \mid u_i)$, $k(v \mid v_j)$ as kernels. Furthermore, because $\IP[a;b]$ and $\Iq[u;v]$ are the mutual information before and after embedding, Lemma 2 implies that $\DKL{P}{Q}$ is bounded by the loss of the mutual information by the embedding.

Lemma 1 and 2 can be extended to the co-occurrence data of three or more elements, the higher-order co-occurrence data.

\begin{lemma}
  Assume three sets $A=\{a_i\}$, $B=\{b_j\}$, and $C=\{c_k\}$, with the empirical co-occurrence probability of $(a_i,b_j,c_k)$ given as $P_{ijk}\equiv P(a_i,b_j,c_k)$. Suppose further that $q(u,v,w)$ is the modeled co-occurrence density in the joint latent space $\cU\times\cV\times\cW$, and $Q(a_i,b_j,c_k)$ is the modeled co-occurrence probability given as
  \begin{align}
      Q(a_i,b_j,c_k)=\iiint P(a_i\mid u)\, P(b_j\mid v)\, P(c_k\mid w)\, q(u,v,w) \du\dv\dw.
      \eqlabel{Qijk}
  \end{align}
  When $U=\{u_i\}$, $V=\{v_j\}$ and $W=\{w_j\}$ are given, the upper bound of the KL divergence between $P(a_i,b_j,c_k)$ and $Q(a_i,b_j,c_k)$ is minimized with respect to $q$, if and only if
  \begin{align*}
      q(u,v,w)=\sum_{i,j,k} k(u\mid u_i)\,k(v\mid v_j)\,k(w\mid w_k)\,P(a_i,b_j,c_k),
  \end{align*}
  and the upper bound of $\DKL{P}{Q}$ is given as
  \begin{align*}
      \DKL{P}{Q}\leq \IP[a;b;c]-\Iq[u;v;w],
  \end{align*}
  where
  \begin{align*}
      \IP[a;b;c]&=\sum_{i,j,k} P(a_i,b_j,c_k)\log\frac{P(a_i,b_j,c_k)}{P(a_i)P(b_j)P(c_k)} \\
      \Iq[u;v;w]&=\iiint p(u,v,w)\log\frac{p(u,v,w)}{p(u)p(v)p(w)}\du\dv\dw.
  \end{align*}
  The same is true for co-occurrence data of fourth order or higher.
\end{lemma}

The proof of Lemma 3 is the same as for Lemma 1 and 2. $\IP[a;b;c]$ and $\Iq[u;v;w]$ are the total correlation of $a,b,c$ and $u,v,w$, i.e., the generalized mutual information. The detailed proof for the three-domain extension is provided in Appendix A.

\subsubsection*{\bf Theoretical significance.}
Lemmas~1 to 3 collectively establish the theoretical core of our proposed method.
Specifically, Lemma~1 shows that minimizing the KL divergence between empirical and model co-occurrence distributions is equivalent to maximizing the retained mutual information after embedding.
Lemma~2 further ensures that the modeled density $q(u,v)$ is uniquely determined in closed form via kernel density estimation, avoiding the need for variational parameters or auxiliary optimization.
Lemma~3 generalizes this principle to higher-order co-occurrence structures using total correlation.
Together, these results provide a principled justification for the proposed objective function and its applicability to multi-domain visual exploration.
In this sense, the lemmas are not auxiliary results, but constitute the analytical foundation that distinguishes our approach from existing heuristic or deep learning-based embeddings.

\subsection{Objective function and optimization}
\subsubsection*{\bf Objective function and optimization method}

Based on Lemma 2, the modeled co-occurrence density $q(u,v)$ is uniquely determined as a kernel density estimator over the embedding vectors $U=\{u_i\}$ and $V=\{v_j\}$.  
Substituting this form into Lemma 1 leads to an upper bound on $\DKL{P}{Q}$ that depends solely on the latent variables $U$ and $V$.  
This allows us to define an objective function over the embedding space as follows:
\begin{align}
    F_\text{obj}[U,V] &:= \Iq[u;v] - \lambda R(U,V) \notag\\
      &= \iint q(u,v\mid U,V) \log \frac{q(u,v\mid U,V)}{q(u)q(v)} 
      \du \dv - \lambda R(U,V) \notag\\
      &\simeq \sum_{i,j} P(a_i,b_j)
        \log \frac{q(u_i,v_j\mid U,V)}{q(u_i)q(v_j)} - \lambda R(U,V),
  \eqlabel{OF} \\
  \text{where }q(u,v\mid U,V) &= \sum_{i,j} k(u \mid u_i)\, k(v \mid v_j)\,P(a_i,b_j). \notag
\end{align}

Unlike conventional embedding approaches, the kernel density $q(u,v)$ is not fixed but dynamically evolves during optimization, as it is defined directly by $U$ and $V$.  
Thus, the gradient $\partial F_\text{obj}/\partial u_i$ includes not only the direct effect via $q(u_i, v_j)$, but also indirect effects through neighboring embeddings, such as $\partial q(u_{i'}, v_j)/\partial u_i$ for $i' \ne i$.  

This leads to a softly coupled potential field over the latent space, where each embedding interacts with others via the shared density.  
The result is a smooth and approximately symmetric optimization landscape, which helps coordinate the embedding vectors and reduces the risk of poor local optima.

The regularization term
\begin{align*}
    R(U,V) := \sum_i \|u_i\|_l^l + \sum_j \|v_j\|_l^l
\end{align*}
is introduced to prevent trivial solutions in which the embeddings diverge in order to maximize mutual information.  
In this work, we use either the $L_2$-norm (i.e., Gaussian prior) or $L_\infty$-norm (i.e, uniform prior over $[-1,+1]^d$), depending on the visualization purpose.

Optimization is performed using steepest gradient descent or its variants, applied directly to the embedding vectors. This approach eliminates the need for separate encoder or decoder networks, as the density $q(u,v)$ is defined in closed-form via kernel density estimation.  
The initialization of $U$ and $V$, and the dynamics it induces, are discussed in later.

\vspace{1em}
\subsubsection*{\bf Cases with three or more domains}
Our framework can be naturally extended to co-occurrence data involving three or more domains.
For example, in the case of three domains (e.g., subject–verb–object), the objective function becomes:
\begin{align}
    F_\text{obj}[U,V,W] &:= \Iq[u;v;w] -\lambda R(U,V,W)\notag\\
      &= \iiint q(u,v,w) \log \frac{q(u,v,w)}{q(u)q(v)q(w)} 
      \du \dv \dw -\lambda R(U,V,W)\notag\\
      &\simeq \sum_{i,j,k} P(a_i,b_j,c_k)
        \log \frac{q(u_i,v_j,w_k)}{q(u_i)q(v_j)q(w_k)}
        -\lambda R(U,V,W).
  \eqlabel{OF2}
\end{align}

For co-occurrence data involving four or more domains, the objective function can be extended in a straightforward manner by generalizing the total correlation and using corresponding kernel density models.
To the best of our knowledge, such direct optimization of higher-order co-occurrence structures has not been addressed in previous work, making this a notable strength of our method.

\subsubsection*{\bf Initialization and diffusion dynamics}

The embedding vectors $U=\{u_i\}$ and $V=\{v_j\}$ are initialized from a small distribution near the origin.  
We consider two initialization strategies: (1) sampling from a Gaussian distribution $\mathcal{N}(0, \sigma^2 I)$, and (2) using principal component analysis (PCA) on the co-occurrence matrix. In both cases, the initial spread is set to be much smaller than the kernel bandwidth—typically less than one-tenth—so that $q(u,v)$ starts with strong smoothing. This results in a natural form of simulated annealing, where both smoothing and stochasticity gradually diminish. 

In case (1), we employ a noisy gradient method to eliminate dependence on the initial state. As learning progresses, the embeddings spread out, and the noise becomes relatively weaker compared to the scale of the latent space. This results in a natural form of simulated annealing, where both smoothing and stochasticity gradually diminish.

In case (2), the PCA initialization provides a reasonably optimized starting point, allowing stable convergence with a standard (deterministic) gradient method. It is worth noting that, in the initial state, the kernel smoothing effect is extremely strong, resulting in a state that is nearly equivalent to linear fitting. Therefore, using PCA as the initialization is theoretically justified.

In both settings, the regularization term eventually prevents excessive diffusion and leads to a stable configuration of the embeddings.

\subsection{\bf Progressive training strategy for stable mutual information optimization}

While the proposed method aims to maximize the mutual information $I_q[u;v]$ between latent variables, a fundamental challenge arises in the early stages of training: the modeled density $q(u,v)$ is initially uninformative due to randomly initialized embeddings. Consequently, $I_q[u;v]$ reflects mostly noise rather than meaningful structure, and naive optimization may easily converge to poor local optima.

To address this issue, we adopt a progressive two-stage training strategy. In the first stage, we optimize auxiliary objectives that provide more stable and interpretable gradient signals:
\[
F_{\mathrm{aux},u} = I[u; b], \quad F_{\mathrm{aux},v} = I[v; a].
\]
These objectives encourage the embeddings $U$ and $V$ to align with the marginal distributions of the observed co-occurrence data, thereby initializing the model in a semantically meaningful region of the latent space. Once the embeddings have acquired sufficient structure, we switch to directly optimizing the main objective $\Iq[u;v]$.

The auxiliary objectives are computed using partially modeled co-occurrence densities:
\[
q_u(u,b_j) = \sum_i P(a_i,b_j) k(u \mid u_i), \quad
q_v(a_i,v) = \sum_j P(a_i,b_j) k(v \mid v_j),
\]
with their marginals:
\[
q(u) = \sum_j q_u(u, b_j), \quad
q(v) = \sum_i q_v(a_i, v).
\]
These yield tractable approximations:
\[
F_{\mathrm{aux},u} \approx \sum_{i,j} P(a_i,b_j) \log \frac{q_u(u_i,b_j)}{q(u_i) P(b_j)}, \quad
F_{\mathrm{aux},v} \approx \sum_{i,j} P(a_i,b_j) \log \frac{q_v(a_i,v_j)}{P(a_i) q(v_j)}.
\]

Importantly, this warm-up strategy is not merely a heuristic but is grounded in theory. The following inequality offers theoretical justification:
\[
I[u; v] \geq I[u; b] + I[a; v] - I[a; b],
\]
showing that increasing $I[u; b]$ and $I[a; v]$ contributes to improving the mutual information between the latent variables, as long as $I[a; b]$ is fixed by the data. While this bound is not always tight, it provides a meaningful theoretical link between the auxiliary and main objectives.

Although the auxiliary objectives do not directly optimize the joint mutual information $I[u;v]$, this separation proves beneficial: it prevents premature coupling between $u$ and $v$ caused by random initialization, allowing the model to capture global structure more reliably. In practice, we find that this warm-up phase alone yields embeddings close to the final solution, with only minor refinement needed afterwards. Thus, the warm-up is not just a practical convenience, but an essential component that ensures robust convergence and minimizes sensitivity to initialization.

\subsection{Incorporating non-co-occurrence for training stability}
\label{non-co-occurrence}

The method described so far has focused exclusively on modeling observed (positive) co-occurrence events. While this formulation is theoretically well-grounded and leads to the same optimal solution under idealized conditions, it may suffer from practical limitations---especially when applied to sparse or incomplete data. In such settings, the absence of co-occurrence can carry informative signals, and ignoring it may hinder both training stability and embedding quality.

We identify two specific issues that can arise when non-co-occurrence is not explicitly modeled:
\begin{itemize}
  \item \textbf{Lack of supervision in sparse regions:}  
  When the empirical co-occurrence probability $P(a_i, b_j)$ is close to zero, the loss function becomes nearly insensitive to how the model estimates $q(u_i, v_j)$. Consequently, large parts of the latent space remain weakly constrained, particularly during the early stages of training. This can lead to unstable or arbitrary embedding configurations.

  \item \textbf{Failure to capture exclusivity:}  
  A low co-occurrence probability does not distinguish between unrelated pairs and those that are semantically incompatible (e.g., antonyms or domain-specific oppositions). Without explicitly modeling the absence of co-occurrence, the learned embeddings may fail to represent mutual exclusivity in a meaningful way.
\end{itemize}

The value of incorporating non-co-occurrence has also been recognized in prior work, such as negative sampling in skip-gram models~\cite{Mikolov2013} and the use of shifted PPMI~\cite{Levy20142177}, where subtracting a constant emphasizes deviations from random co-occurrence. These approaches suggest that even non-events can provide informative constraints on the embedding space.

\subsubsection*{\bf Generative model of co-occurrence events}
To incorporate both co-occurrence and non-co-occurrence information, we introduce the following generative process.

Let \( a_i \in A \) and \( b_j \in B \) be elements drawn independently from two sets, according to the marginal probabilities \( P(a_i) = P_i \) and \( P(b_j) = P_j \), respectively. Thus, the probability of sampling the pair \( (a_i, b_j) \) is \( P_i \cdot P_j \). Given a sampled pair, we assume that it is accepted (i.e., observed as a co-occurrence) with probability \( P(c = 1 \mid a_i, b_j) \), and rejected (i.e., treated as non-co-occurrence) with probability \( P(c = 0 \mid a_i, b_j) \). In this setup, a co-occurrence (positive event) is observed only when the pair is accepted.

Under this generative model, the joint distribution over \( (c, a_i, b_j) \) is given by:
\[
P(c, a_i, b_j) = P(c \mid a_i, b_j) \cdot P(a_i) \cdot P(b_j).
\]

The conventional co-occurrence probability corresponds to the conditional probability of a pair given a positive event:
\[
P(a_i, b_j) = P(a_i, b_j \mid c = 1).
\]
Accordingly, the relationship between the joint distribution and the conventional co-occurrence probability is:
\[
P(c = 1, a_i, b_j) = P(a_i, b_j \mid c = 1) \cdot P(c = 1).
\]

\subsubsection*{\bf KL divergence and its upper bound}

We define a probabilistic model that incorporates the binary co-occurrence variable \( c \in \{0, 1\} \) as \( q(c, u, v) \), where \( u \) and \( v \) denote latent representations of \( a_i \in A \) and \( b_j \in B \), respectively.

Let \( Q(c, a_i, b_j) \) be the model-induced joint probability over the observed and latent variables. It is computed as:
\[
Q(c, a_i, b_j) = \iint P(a_i \mid u)\, P(b_j \mid v)\, q(c, u, v)\, du\, dv.
\]
Assuming kernel-based decoding, this can be rewritten as:
\[
Q(c, a_i, b_j) = P_i P_j \iint k(u \mid u_i)\, k(v \mid v_j)\, \frac{q(c, u, v)}{q(u)\, q(v)}\, du\, dv,
\]
where \( k(u \mid u_i) \) and \( k(v \mid v_j) \) are kernel functions centered at \( u_i \) and \( v_j \), and \( q(u) \), \( q(v) \) are the corresponding marginal densities.

Under this formulation, the following lemma provides an upper bound on the KL divergence between the empirical and model distributions.

\begin{lemma}
Let \( P(c, a_i, b_j) \) denote the empirical joint distribution over co-/non-co-occurrence events and item pairs, and let \( Q(c, a_i, b_j) \) be the model-induced distribution as defined above. Then, the KL divergence is upper-bounded as follows:
\[
D_\mathrm{KL}[P \| Q] \leq I_{P}[c; a; b] - I_{q}[c; u; v],
\]
where the model distribution in latent space is defined by
\[
q(c, u, v) = \sum_{i,j} P(c, a_i, b_j)\, k(u \mid u_i)\, k(v \mid v_j),
\]
and \( I_{q}[c; u; v] \) denotes the mutual information between \( c \), \( u \), and \( v \) under \( q \).
\end{lemma}

This result generalizes Lemma~1 by explicitly including the binary variable \( c \), showing that the mutual-information–based objective remains theoretically valid even when non-co-occurrence events are incorporated. (See Appendix~A for proof.)

The mutual information term \( I_{q}[c; u; v] \) in the upper bound can be further decomposed as:
\begin{align*}
I_{q}[c; u; v] 
&= \sum_{c} \iint q(c, u, v) \log \frac{q(c, u, v)}{P(c)\, q(u)\, q(v)} \, du\, dv \\
&= \sum_{c} P(c)\, I[u; v \mid c] + I[c; u] + I[c; v],
\end{align*}
where \( I[u; v \mid c] \) denotes the conditional mutual information between \( u \) and \( v \) given \( c \), and \( I[c; u] \), \( I[c; v] \) capture dependencies between \( c \) and each latent variable.

Notably, the term for \( c = 1 \) corresponds to the original objective in the positive-only formulation. By incorporating the \( c = 0 \) component, this extended objective improves training dynamics in several ways: it promotes disentanglement via \( I[c; u] \), enhances stability in sparse regions through \( I[u; v \mid c = 0] \), and maintains consistency with the positive-only solution under mild conditions.

\subsubsection*{\bf Estimating co-/non-co-occurrence probability}

In practice, however, only co-occurring pairs ($c=1$) are observed. The probability of non-co-occurrence, $P(c=0 \mid a_i, b_j)$, is not directly observable—placing us in a standard \emph{positive-unlabeled (PU) learning} scenario. To estimate $P(c \mid a_i, b_j)$, we adopt a class-prior–adjusted estimator:
\begin{align*}
P(c = 1 \mid a_i, b_j) &= \frac{N_1(a_i, b_j) + \alpha}{N_1(a_i, b_j) + N_0(a_i, b_j) + \alpha + \beta}, \\
P(c = 0 \mid a_i, b_j) &= 1 - P(c = 1 \mid a_i, b_j),
\end{align*}
where $N_1(a_i, b_j)$ is the observed co-occurrence count, and $N_0(a_i, b_j)$ is estimated as:
\[
N_0(a_i, b_j) = \frac{\beta}{\alpha} \, N_1 \, P(a_i \mid c = 1) \, P(b_j \mid c = 1).
\]
Here, $N_1$ is the total number of observed co-occurrence events, and $\alpha$, $\beta$ are hyperparameters representing the expected ratio of co-occurrence to non-co-occurrence events. This estimation method follows standard PU learning techniques~\cite{Elkan2008213,Bekker2020719}. Details of this estimation procedure, including prior assumptions and smoothing, are provided in Appendix~B.

In the remainder of this paper, we adopt the extended formulation with non-co-occurrence as the default. 
This formulation improves training stability by providing informative gradients in sparse regions, while preserving the ability to recover embeddings consistent with the positive-only formulation under mild assumptions. 
Accordingly, all subsequent experiments are conducted using this version.

\subsection{Estimating indirect co-occurrence via Markov diffusion}
\label{sec:sparse}

Real-world co-occurrence data are often extremely sparse in practice---especially in higher-order settings such as subject--verb--object triples. To alleviate this, we estimate \emph{indirect co-occurrence} using a Markov diffusion process over the bipartite co-occurrence graph:
\[
P^{(2)}(a_i, b_j \mid c=1) = \sum_{i', j'} P(b_j \mid a_{i'}, c=1) \, P(a_{i'} \mid b_{j'}, c=1) \, P(b_{j'} \mid a_i, c=1) \, P(a_i).
\]
This second-order expansion corresponds to a random walk $a_i \to b_{j'} \to a_{i'} \to b_j$, enabling the estimation of plausible associations even for item pairs with no direct co-occurrence. The idea naturally generalizes to $P^{(m)}(a_i, b_j \mid c=1)$ via $m$-step alternating walks, offering flexible control over the diffusion depth.

Graph-theoretically, this corresponds to applying powers of the biadjacency matrix to extract higher-order neighborhood structure. While diffusion-based smoothing is a standard technique, our integration into a \emph{mutual information framework over dual latent spaces} is, to the best of our knowledge, both novel and essential for robust modeling under extreme sparsity.

\section{Interactive visual exploration methods}

\subsection{Intra-domain and inter-domain analysis}

Heterogeneous co-occurrence data involve two distinct types of relationships: \emph{intra-domain} relationships within the same domain (e.g., adjective--adjective or noun--noun), and \emph{inter-domain} relationships across different domains (e.g., adjective--noun). Intra-domain analysis focuses on the similarity or clustering of elements within a domain, such as the proximity between ``quick'' and ``lazy'' or between ``fox'' and ``dog''. In contrast, inter-domain analysis examines how elements from different domains co-occur---for instance, how the adjective ``quick'' relates to the noun ``fox''. Our method accommodates both types of relationships in a unified probabilistic embedding framework, enabling seamless analysis across and within domains.

\subsection{Interactive visualization for inter-domain analysis}

In visualizing heterogeneous co-occurrence data, two distinct structures must be considered: intra-domain organization and inter-domain associations. While intra-domain structure---such as the spatial distribution of adjectives or nouns---can be effectively visualized through 2D scatter maps of latent embeddings, inter-domain relationships are far more challenging. These relationships span across domains (e.g., adjective--noun), often exhibit asymmetry, and cannot be easily captured in a single geometric space. Visualizing such cross-domain interactions in an interpretable and flexible manner remains a central challenge.

Existing visualization methods often struggle with this challenge. A representative example is CODE~\cite{Globerson20072265}, which embeds all elements---regardless of domain---into a shared latent space, where proximity reflects co-occurrence frequency. While effective for capturing overall structure, this approach enforces symmetry on relationships and merges all domains into a single geometry. As a result, it becomes difficult to assess directional influence, such as whether a noun tends to attract certain adjectives or vice versa. Furthermore, once such visualizations are generated, they are static: analysts can observe what is already rendered, but cannot pose new questions or obtain dynamic, conditional views. 

This limitation is visually evident in seriation-based heatmaps (Figure~\ref{fig:seriation}), where the co-occurrence pattern often exhibits strong asymmetry across domains. Symmetric embeddings such as CODE can be interpreted as focusing primarily on the diagonal structure of such matrices, effectively ignoring directional variation and marginal asymmetries that may carry semantic significance.

To address these limitations, we propose an interactive framework that explicitly models the joint co-occurrence density $q(u,v)$, where $u$ and $v$ are latent variables in separate embedding spaces $\mathcal{U}$ and $\mathcal{V}$ corresponding to different domains. This formulation enables the computation of conditional distributions $q(v \mid u)$ and $q(u \mid v)$, allowing directional and asymmetric analysis across domains.

Building on this, we introduce a technique called \emph{coloring-by-conditional-probability} (CbCP), which allows analysts to explore cross-domain relationships interactively. When a user selects a specific element $v^* \in \mathcal{V}$ (e.g., a particular adjective), it serves as the \emph{target of interest} (ToI), and the embedding space $\mathcal{U}$ (e.g., nouns) is colored with a heatmap based on the conditional distribution $q(u \mid v^*) = q(u, v^*) / q(v^*)$. Conversely, selecting $u^* \in \mathcal{U}$ as the ToI triggers a heatmap over $\mathcal{V}$ using $q(v \mid u^*) = q(u^*, v) / q(u^*)$. This process can be repeated, enabling users to alternate between domains and follow a guided path through asymmetric associations.

This iterative interaction can be viewed as a form of Markov exploration, in which each step involves conditioning on a new ToI to update the distribution over the opposite domain. In symmetric models such as CODE, such transitions yield little variation—akin to Brownian motion in place—because the embedding space is shared and undirected. In contrast, our framework leverages asymmetric dependencies, enabling long-range semantic transitions. These include “jumps” to semantically distant regions in the latent space, such as off-diagonal hotspots observed in seriation heatmaps (Figure~\ref{fig:seriation}). This mechanism supports what we refer to as \emph{semantic teleportation}.

Such a capability is especially valuable when the domains differ in granularity—for example, when adjectives exhibit fine-grained variation and nouns provide broader contextual structure. By supporting asymmetric, bidirectional queries centered around a changing ToI, CbCP enables analysts to conduct structured, interpretable exploration of heterogeneous co-occurrence data within a coherent probabilistic framework.

\subsection{Extension to three or more domains}

CbCP naturally extends to datasets involving three or more domains, where inter-domain relationships become even more complex and informative. Consider subject--verb--object (SVO) co-occurrence data. Even when the subject remains fixed (e.g., \textit{child}), its co-occurring objects may vary drastically depending on the verb (e.g., \textit{learn} vs. \textit{love}). In such cases, analyzing pairwise relations is insufficient; instead, higher-order conditional structures must be examined.

When a single target of interest (ToI) is specified---for instance, fixing $u^* \in \mathcal{U}$ (e.g., subject = \textit{child})---CbCP can visualize conditional distributions over multiple other domains, such as $q(v \mid u^*)$ and $q(w \mid u^*)$, showing the verbs and objects that tend to co-occur with the given subject. This enables exploration of the ToI's contextual behavior across domains.

Furthermore, when two targets are specified---e.g., $u^*$ and $v^*$ corresponding to a subject and a verb---we can condition jointly and visualize $q(w \mid u^*, v^*)$ over the object domain. For example, setting ToI = (\textit{child}, \textit{learn}) may highlight objects such as \textit{knowledge}, \textit{language}, or \textit{skill}, revealing semantically coherent triples. These multi-way queries preserve domain separation while supporting rich, asymmetric exploration across higher-order co-occurrence structures.

Taken together, our approach offers capabilities that are rarely found in existing systems.  
To our knowledge, few prior systems simultaneously satisfy the following three properties:
\begin{itemize}
  \item maintaining separate latent spaces for each domain,
  \item modeling full asymmetric co-occurrence distributions, and
  \item supporting dynamic conditional queries across two or more domains.
\end{itemize}
The proposed CbCP-based interface offers a simple yet general mechanism to support these capabilities within a unified probabilistic framework.

\section{Experimental results}

We conducted a series of experiments to evaluate the effectiveness of the proposed method on both synthetic and real-world co-occurrence data. Each experiment is designed to highlight different aspects of our approach---such as its ability to capture asymmetric structure, preserve semantic coherence, and support exploratory visualization. All experiments follow a consistent setup, including data preprocessing, dimensionality settings, and optimization procedures. Full details of the experimental configuration are provided in Appendix~B.

\begin{figure}[t]
  \centering
  \begin{minipage}[c]{0.3\linewidth}
    \centering
    \includegraphics[width=\linewidth]{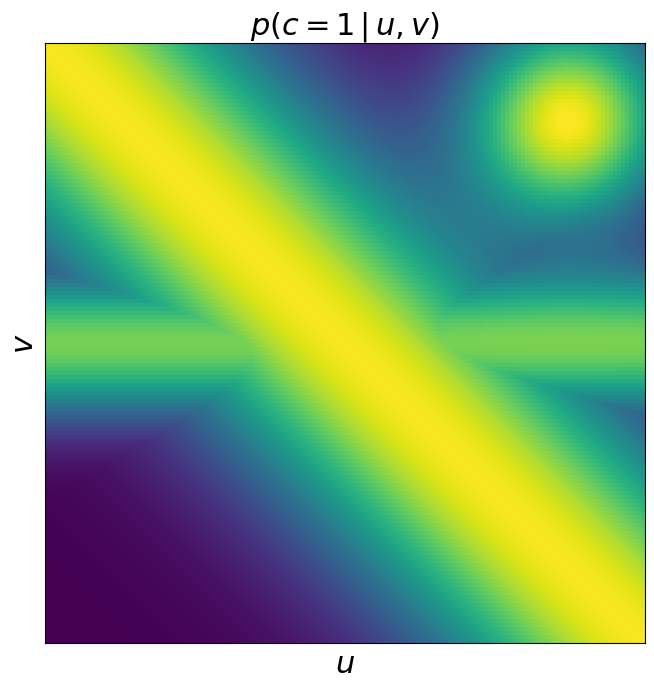}\\
    (a)
  \end{minipage}
  \begin{minipage}[c]{0.3\linewidth}
    \centering
    \includegraphics[width=\linewidth]{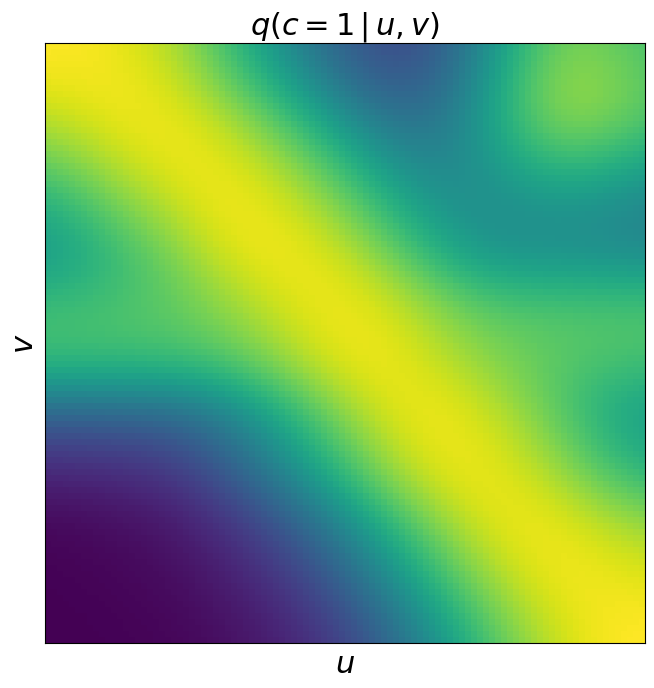}\\
    (b)
  \end{minipage}
  \begin{minipage}[c]{0.3\linewidth}
    \centering
    \includegraphics[width=\linewidth]{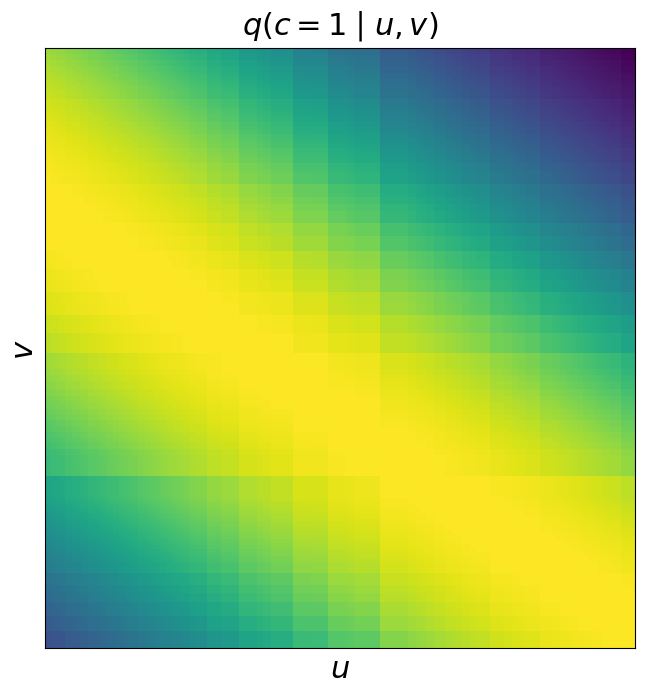}\\
    (c)
  \end{minipage}
  \caption{Results on artificial data. The latent joint probability $p(u, v)$ was synthetically generated, and co-occurrence frequencies were sampled from it.  
(a) The original $p(u, v)$ (ground truth);  
(b) The joint probability $q(u, v)$ modeled by the proposed method;  
(c) The joint probability $q(u, v)$ modeled by CODE.  
These results demonstrate that the proposed method successfully recovers asymmetric structures in the latent space by explicitly modeling the joint distribution $q(u, v)$.
}
  \figlabel{artificial}
\end{figure}

\subsection{Artificial data}

To qualitatively assess whether our method can recover asymmetric structures, we conducted experiments using synthetic co-occurrence data designed to mimic typical patterns observed in real heterogeneous settings. Specifically, we generated co-occurrence data between two distinct sets $A$ and $B$, where the ground-truth distribution $P(a_i, b_j)$ exhibits three characteristic structures: (1) a diagonal ridge representing symmetric alignment (e.g., topical correspondence), (2) a localized off-diagonal spot indicating asymmetric association between specific subsets, and (3) a broad horizontal band reflecting items in $B$ that co-occur widely across elements in $A$ (\figref{artificial}~(a)). This setup reflects essential patterns often found in real-world document--term co-occurrence matrices (\figref{seriation}).

We then applied our proposed method to estimate the joint density $q(u,v)$ from the synthetic data. As shown in \figref{artificial}~(b), the resulting heatmap successfully recovers all three structural components: the diagonal ridge, the off-diagonal asymmetry, and the broad horizontal co-occurrence. While the estimated distribution appears smoother due to kernel density estimation from a finite number of samples, the key asymmetric features are clearly preserved in the learned embedding.

For comparison, we also applied CODE, a representative co-embedding method. As illustrated in \figref{artificial}~(c), the resulting embedding captures the dominant diagonal structure but fails to recover the off-diagonal and horizontal patterns. This outcome reflects a general limitation of conventional methods that do not explicitly model asymmetric co-occurrence relationships.

\begin{figure}[t]
  \centering
  \includegraphics[width=\linewidth]{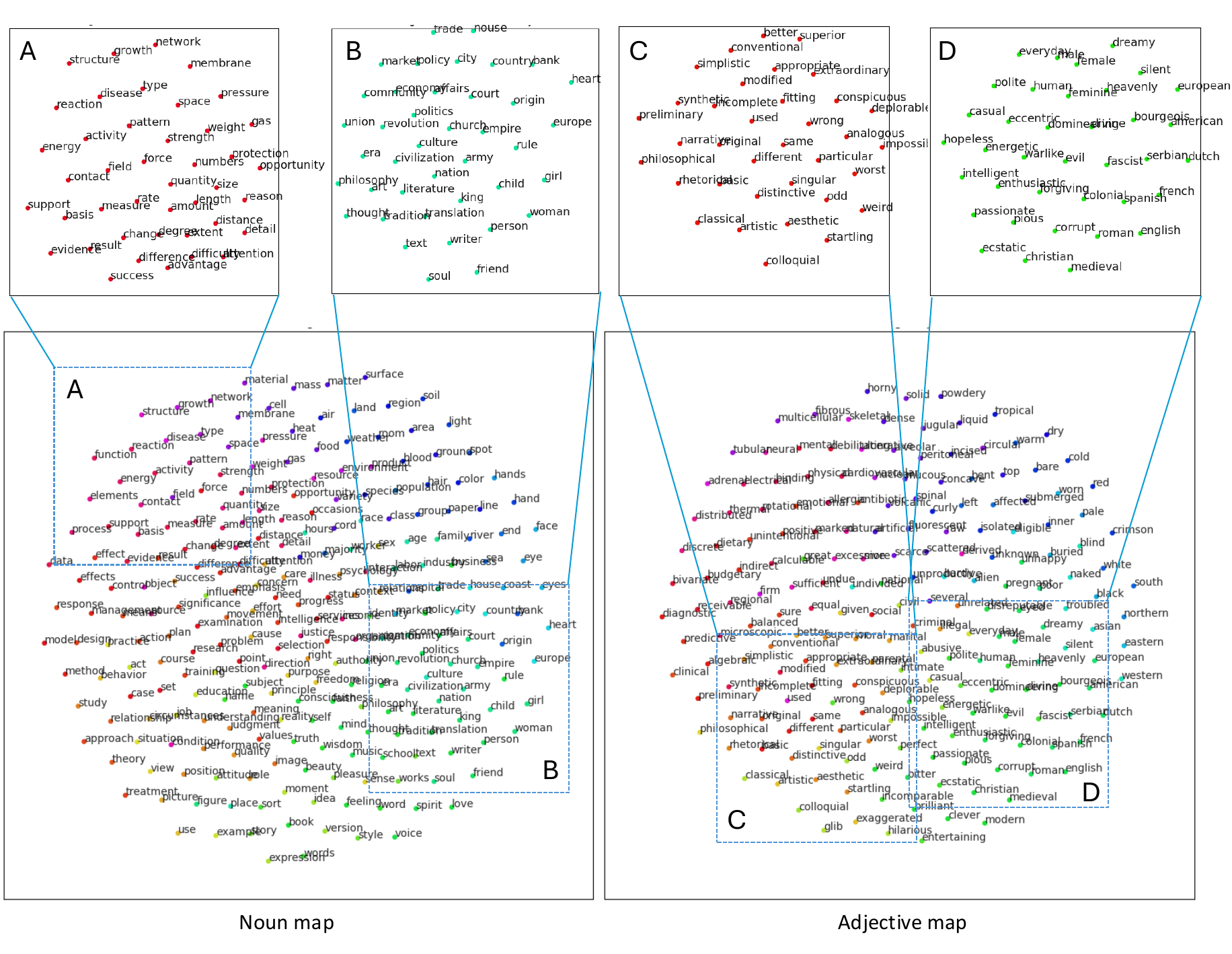}
  \caption{Embedding results for the adjective--noun co-occurrence data. Left: noun map; right: adjective map. Subplots A--D (top) show magnified views of corresponding regions.}
  \figlabel{noun-adj-map}
\end{figure}

\subsection{Adjective--Noun Co-occurrence Data}

To assess the applicability of our method to real-world heterogeneous data, we applied it to a dataset of adjective--noun co-occurrence pairs, where each pair reflects the co-occurrence of an adjective and a noun within the same sentence~\cite{grewal20202196}. This setup inherently involves asymmetry---adjectives modify nouns, but not vice versa---which conventional co-embedding methods struggle to represent due to their use of a shared embedding space. In contrast, our method embeds each domain into a separate latent space, enabling it to preserve both intra-domain semantic coherence and inter-domain asymmetry. The following analyses demonstrate that the resulting embeddings capture meaningful structure within each domain, and reveal directional associations across domains through conditional co-occurrence visualization. \figref{noun-adj-map} provides an overview of the embedding results, with the noun and adjective maps displayed side by side.

\subsubsection*{\bf Intra-domain analysis}

To investigate whether the proposed method captures coherent structure within each domain, we visualized the learned embeddings of nouns and adjectives separately. \figref{noun-adj-map} presents two-dimensional projections: the noun map on the left, and the adjective map on the right.

In the noun map (left), semantically related words tend to cluster together. Region~A (top-left) contains terms associated with physical or scientific concepts such as \emph{force}, \emph{field}, \emph{strength}, and \emph{space}, while Region~B (bottom-right) comprises sociopolitical terms such as \emph{army}, \emph{civilization}, \emph{nation}, and \emph{empire}.

The adjective map (right) exhibits a similar pattern. Region~C (bottom-left) contains abstract relational adjectives such as \emph{different}, \emph{same}, \emph{singular}, and \emph{distinctive}, while Region~D (bottom-right) is populated by adjectives with sociopolitical or ideological meaning, including \emph{colonial}, \emph{fascist}, and \emph{bourgeois}.

These observations suggest that the embeddings reflect latent structure within each domain, based solely on cross-domain co-occurrence patterns. Notably, no direct co-occurrence information within the same domain is used during training. The observed clustering implies that words with similar cross-domain co-occurrence tendencies are located nearby in the embedding space. While some semantic categories---such as sociopolitical concepts---appear across both domains, others (e.g., relational adjectives) are domain-specific.

\begin{figure}[tp]
  \centering
  \includegraphics[width=\linewidth]{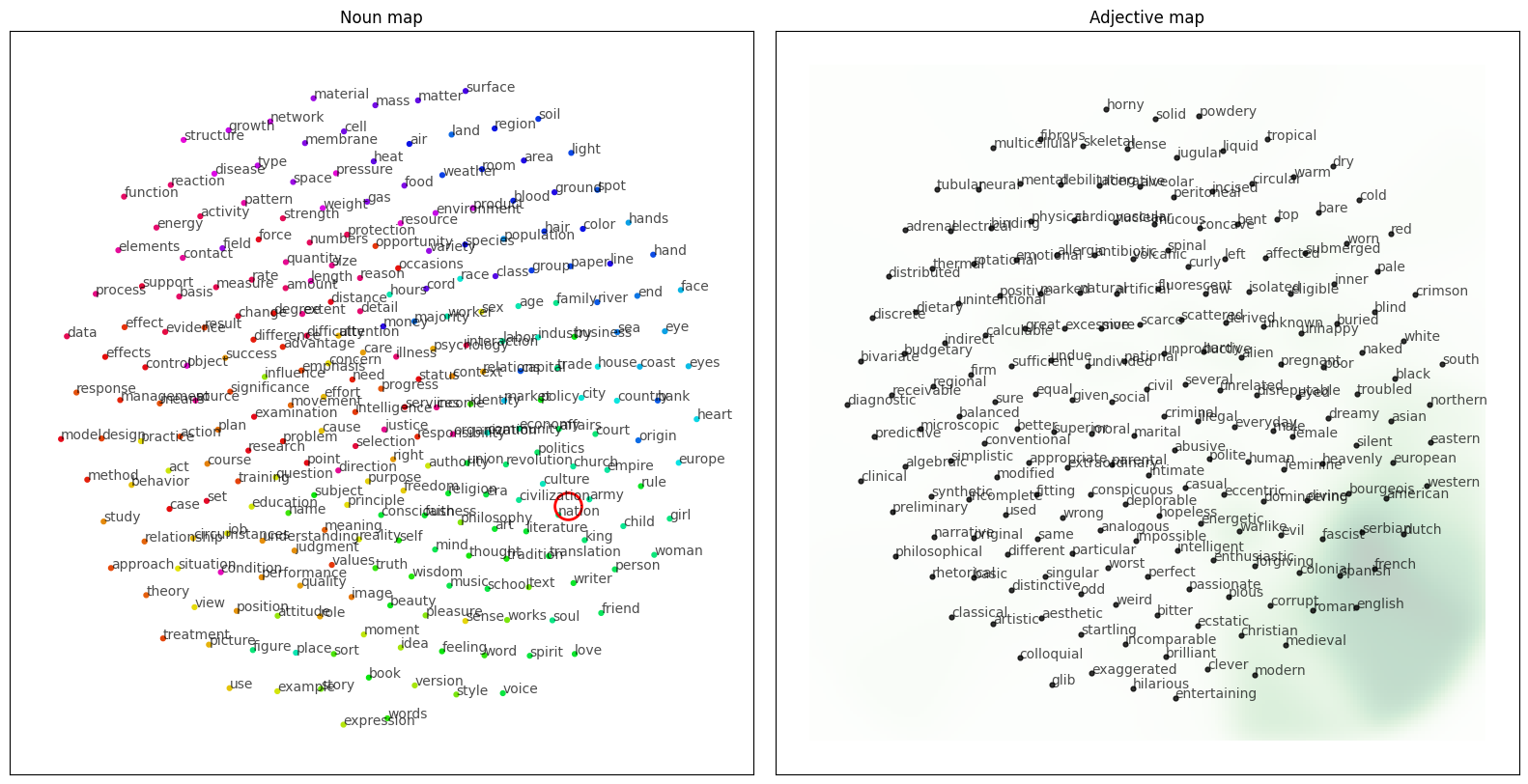}\\
  (a)\vspace{3mm}
  \includegraphics[width=\linewidth]{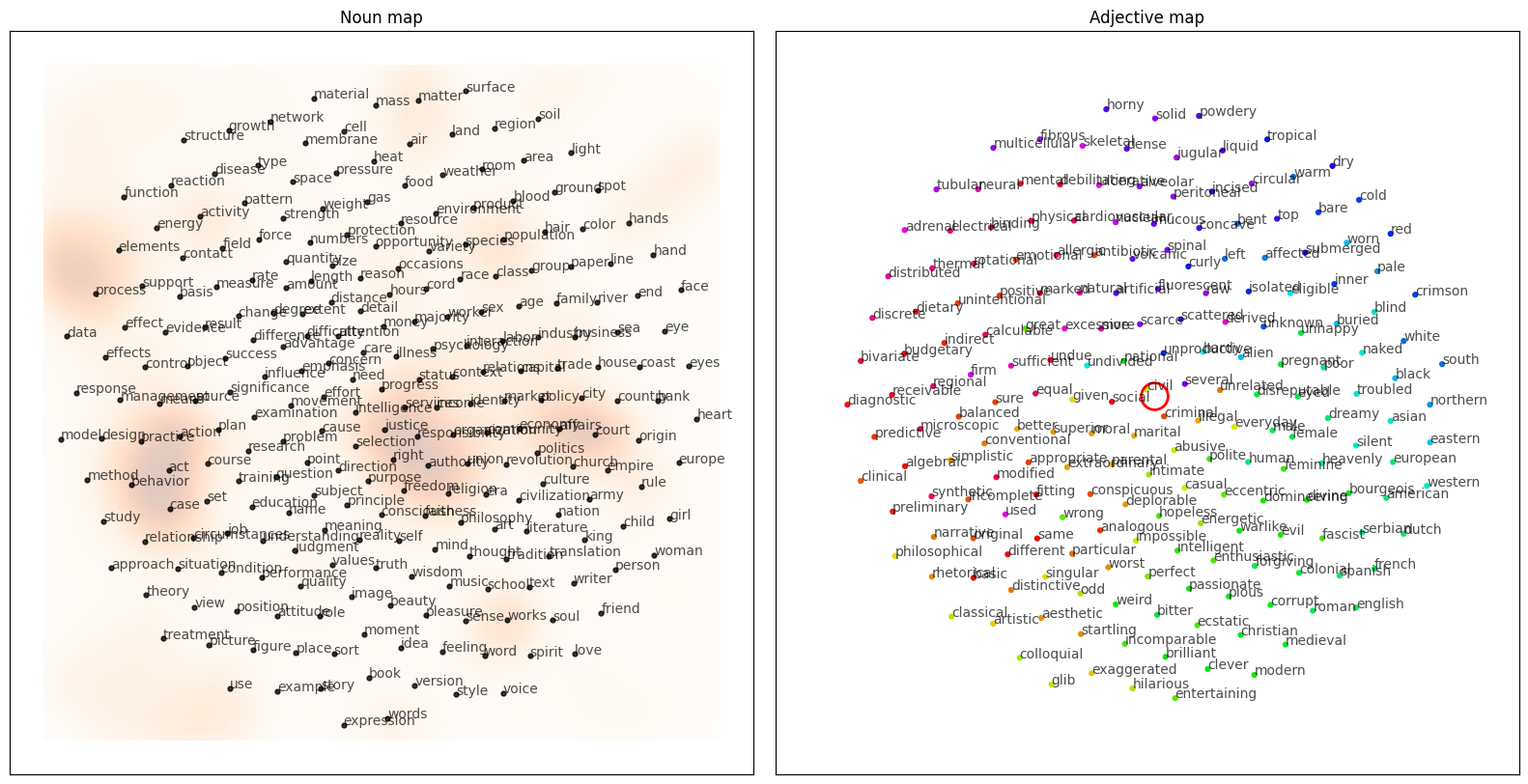}\\
  (b)
  \caption{Visualization of asymmetric co-occurrence via Coloring-by-Conditional-Probability (CbCP). Left: noun map; right: adjective map. (a) When the noun \emph{nation} is selected as the target of interest (red circle), the adjective map highlights a focused set of co-occurring words. (b) Selecting the adjective \emph{civil} reveals multiple noun clusters, each representing a distinct semantic usage.}
  \figlabel{noun-adj-CbCP}
\end{figure}

\subsubsection*{\bf Inter-domain analysis}

To further evaluate how well the model captures asymmetric inter-domain relationships, we used the Coloring-by-Conditional-Probability (CbCP) visualization technique introduced in Section~4. CbCP highlights conditional co-occurrence distributions by overlaying color gradients on one domain’s map based on the conditional probability derived from a target word in the other domain. When a noun $u^*$ is selected as the target of interest (ToI), the conditional distribution $q(v \mid u^*)$ is displayed as a heatmap over the adjective map. Conversely, selecting an adjective $v^*$ allows visualization of $q(u \mid v^*)$ on the noun map.

In our interactive interface, users specify a ToI by clicking on a word in either map. The selected word is marked with a red circle, and the corresponding conditional distribution is rendered as a heatmap over the opposite domain's map, where darker regions indicate higher conditional probability.

\figref{noun-adj-CbCP}(a) shows an example where the noun \emph{nation} is selected as the ToI{}. The adjective map highlights a localized region in the lower-right area, where words such as \emph{French}, \emph{English}, \emph{Spanish}, and \emph{Roman} are concentrated. These adjectives reflect cultural or national identity, and their association with \emph{nation} aligns well with their semantic roles. Adjectives such as \emph{colonial} and \emph{fascist} further emphasize the sociopolitical connotations of the region. This focused pattern suggests a coherent directional relationship between the noun and its modifiers.

\figref{noun-adj-CbCP}(b) shows the reverse direction: the adjective \emph{civil} is selected as the ToI{}. The noun map reveals multiple distinct peaks. One cluster (center) includes terms such as \emph{justice}, \emph{responsibility}, and \emph{right}, corresponding to legal or civic themes. Another cluster (left) contains \emph{act} and \emph{behavior}, reflecting social conduct or action-related contexts. These clearly separated clusters indicate that the adjective \emph{civil} is used in diverse semantic contexts, which the model successfully differentiates in the latent space.

These results demonstrate a key strength of our method: by embedding heterogeneous domains in separate spaces and modeling directional co-occurrence probabilities, it enables the interpretation of asymmetric, many-to-many relationships---a capacity not easily achieved by conventional symmetric co-embedding methods.

\begin{figure}[tp]
  \centering
  \includegraphics[width=\linewidth]{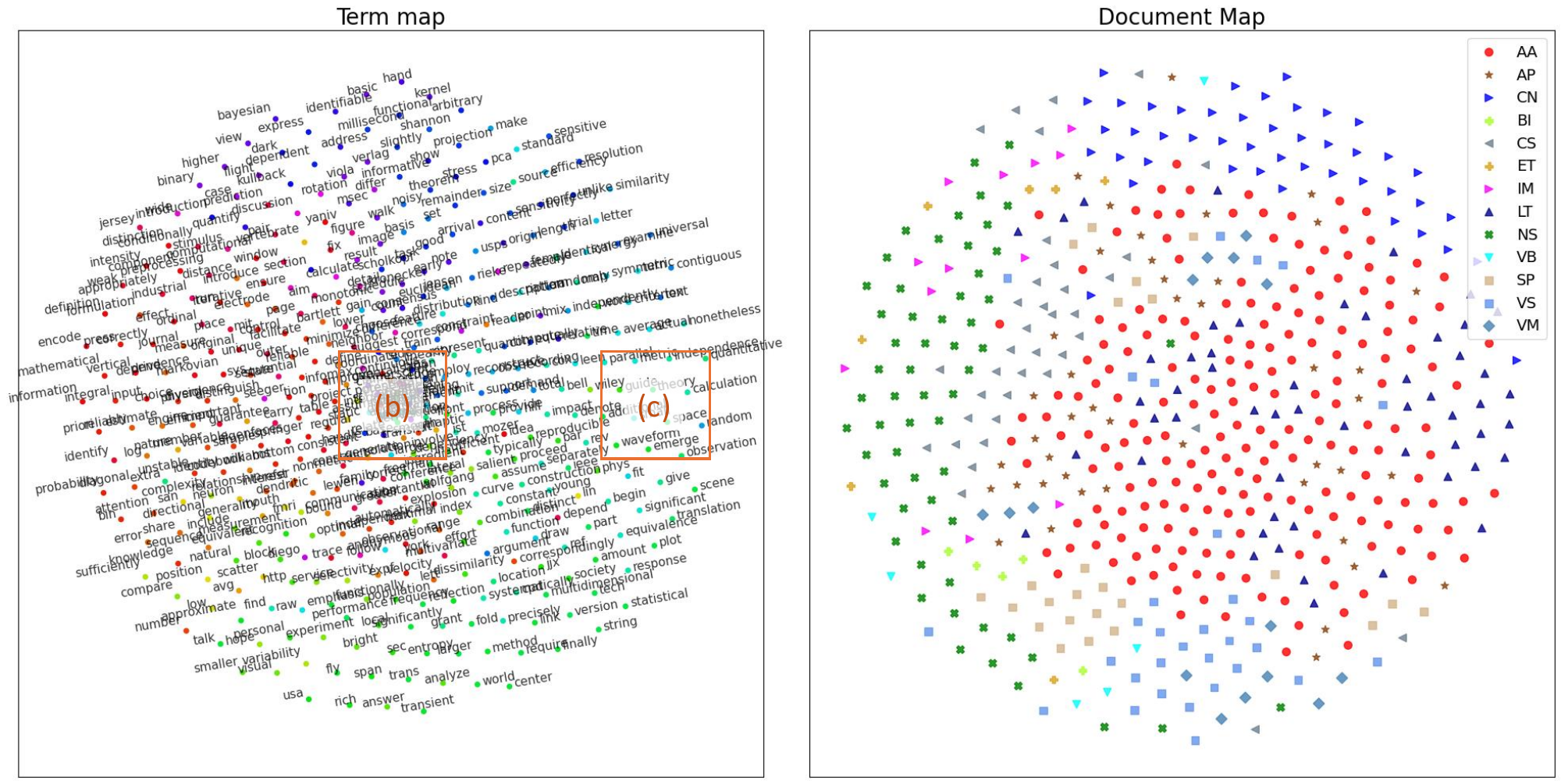}\\
  (a)\\[3mm]
  \begin{minipage}[c]{0.48\linewidth}
    \centering
    \includegraphics[width=0.95\linewidth]{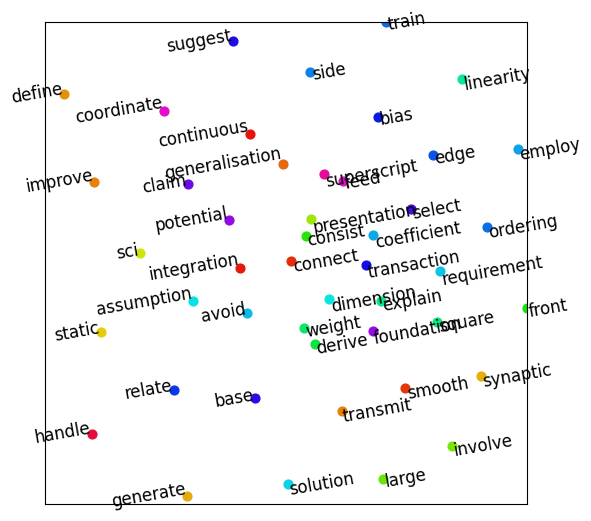}\\
    (b)
  \end{minipage}
  \begin{minipage}[c]{0.48\linewidth}
    \centering
    \includegraphics[width=1.1\linewidth]{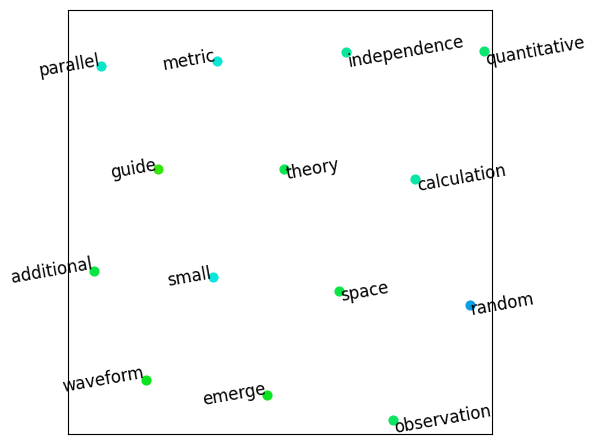}\\
    (c)
  \end{minipage}
  \caption{Embedding results for the NeurIPS document--term co-occurrence data.  
(a) Left: term map; right: document map.  
(b), (c): Zoomed-in views of different regions of the term map.  
Document map markers indicate topic categories as follows:  
AA: Algorithms \& Architectures, AP: Applications, CN: Control \& Reinforcement Learning,  
BI: Brain Imaging, CS: Cognitive Science \& AI, ET: Emerging Technologies, IM: Implementations,  
LT: Learning Theory, NS: Neuroscience, VB: Vision (Biological), SP: Speech and Signal Processing,  
VS: Vision, VM: Vision (Machine).}
  \figlabel{neurips1}
\end{figure}

\begin{figure}[tp]
  \centering
  \includegraphics[width=\linewidth]{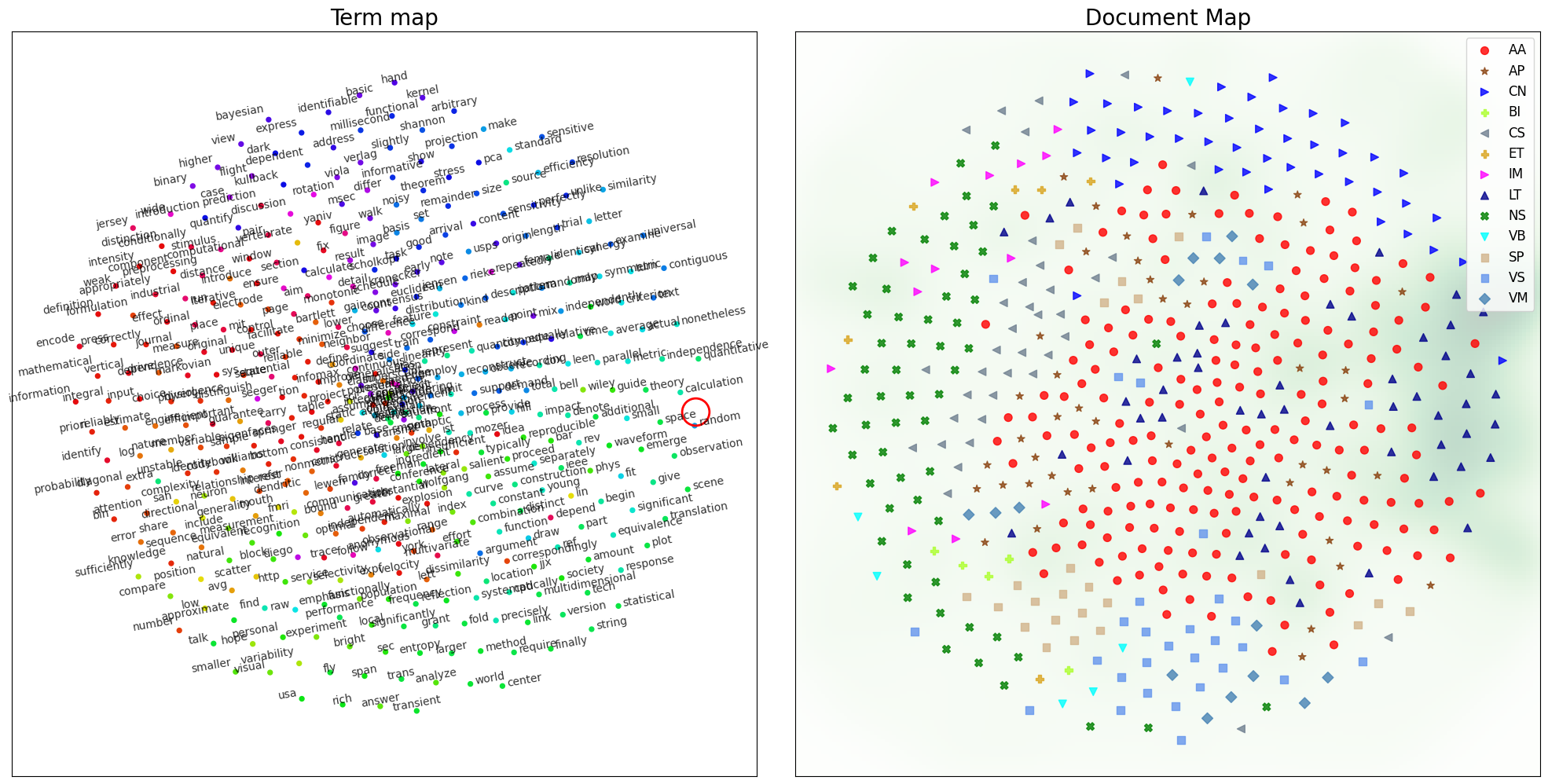}\\
  (a)\\[3mm]
  \includegraphics[width=\linewidth]{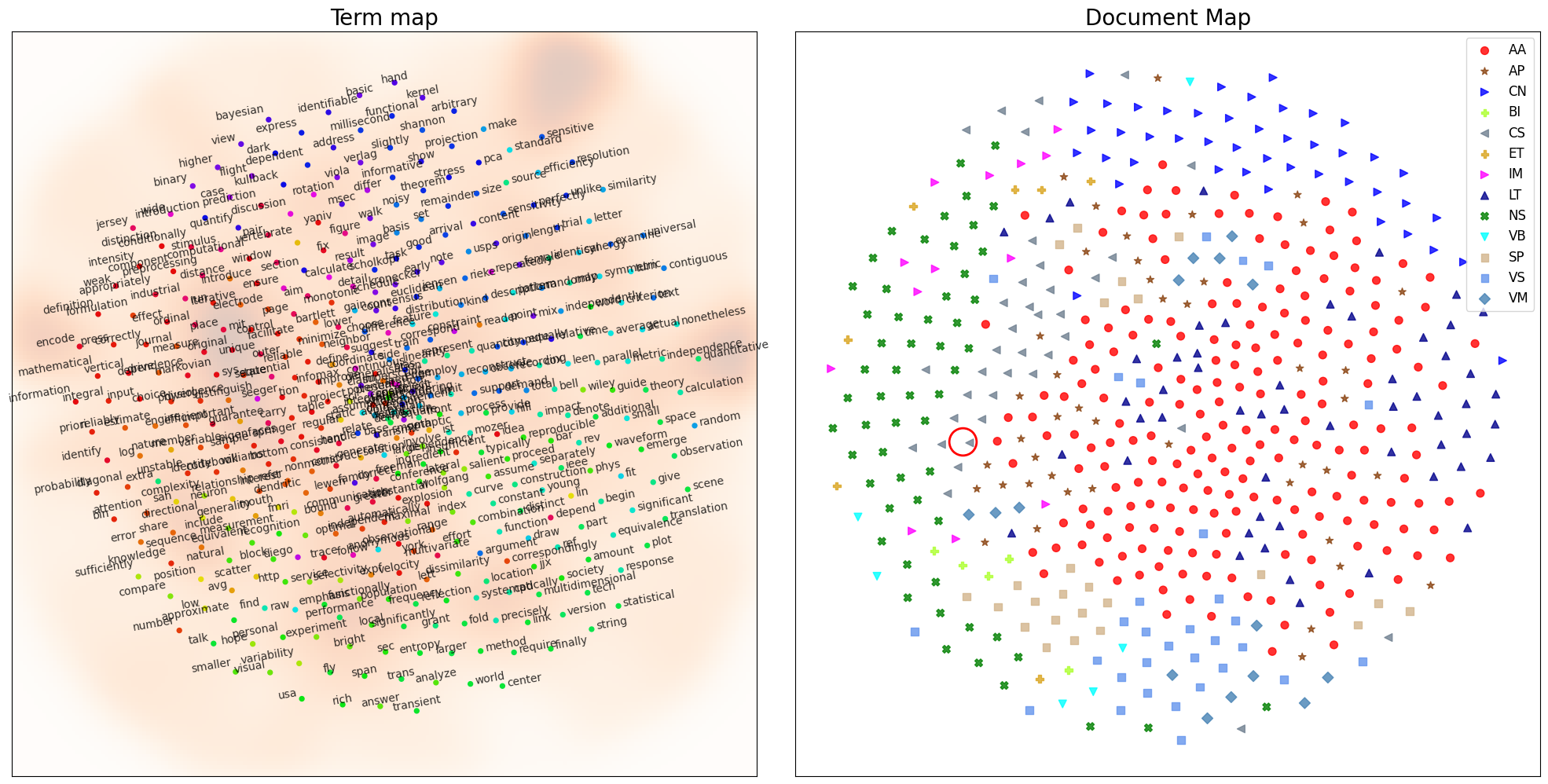}
  (b)
\caption{Conditional co-occurrence visualization using the CbCP technique.  
(a) The red circle in the term map (left) indicates the target of interest (ToI), with terms such as \emph{space} and \emph{random}; the corresponding conditional distribution is visualized on the document map (right).  
(b) The red circle in the document map (right) indicates a ToI from the ``Cognitive Science \& AI'' region; the resulting conditional distribution is shown on the term map (left).}
  \figlabel{neurips2}  
\end{figure}

\subsection{Application to document--term co-occurrence data}

To evaluate the practical utility of the proposed method, we applied it to a document--term co-occurrence dataset constructed from NeurIPS conference papers published between 2001 and 2003.\footnote{https://cs.nyu.edu/~roweis/data.html} We used a subset consisting of 540 papers and 436 words, selected from the full corpus. Empirical co-occurrence probabilities were computed using bag-of-words frequencies. The goal of this experiment is to assess whether the proposed embedding approach supports bidirectional exploration between documents and keywords. \figref{neurips1} shows the resulting embeddings of both domains.

The document embedding (\figref{neurips1}, right) reveals clear clustering patterns that align with topical categories. For instance, papers labeled as ``Vision'' and ``Machine Vision'' appear close together, reflecting their thematic proximity. Similarly, papers in ``Neuroscience'' are positioned near those in ``Cognitive Science and AI,'' suggesting that conceptual overlap between these areas is preserved in the embedding.

The word embedding (\figref{neurips1}, left) also exhibits semantically coherent groupings. In the central region (\figref{neurips1}~(b)), we find general-purpose terms such as \emph{potential}, \emph{generalization}, and \emph{assumption}, which are broadly used across different topics. These words likely correspond to the horizontal bands observed in the seriation plot (\figref{seriation}). In contrast, the right region (\figref{neurips1}~(c)) includes topic-specific terms such as \emph{theory}, \emph{random}, and \emph{metric}, which are characteristic of theoretical machine learning literature.

To investigate cross-domain associations, we employed the CbCP technique introduced earlier. \figref{neurips2} shows the conditional probability heatmaps obtained from representative selections in each domain. In \figref{neurips2}~(a), a target word (ToI, indicated by a red circle) such as \emph{random} or \emph{space} is selected from the theoretical region of the word map. The document map highlights a concentrated region on the right, corresponding to the ``Learning Theory'' category. This reflects a one-to-many, asymmetric association between topic-specific theoretical terms and a coherent cluster of documents.

In the reverse direction (\figref{neurips2}~(b)), a document from the ``Cognitive Science and AI'' region is selected as the ToI. The resulting heatmap on the word map exhibits multiple peaks. One cluster highlights reinforcement learning and control-related terms, while another overlaps with theoretical vocabulary. This dispersion suggests that the selected paper bridges multiple semantic regions. The ability to identify such branching structures enables users to follow topic transitions---for instance, from cognitive science to learning theory---by iteratively selecting nearby peaks as new ToIs. This is a typical case of \emph{semantic teleportation}, where the model reveals distant but meaningful connections through asymmetric co-occurrence.

These findings illustrate the strength of our method in supporting exploratory analysis of complex, many-to-many, and asymmetric co-occurrence relationships in real-world document collections.

\begin{figure}[tp]
  \centering
  \begin{minipage}[c]{0.48\linewidth}
    \centering
    \includegraphics[width=\linewidth]{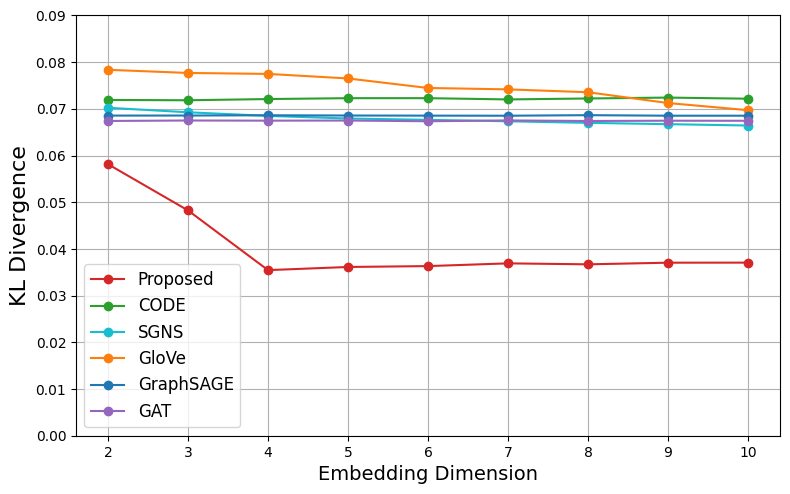}\\
    (a)
  \end{minipage} 
  \begin{minipage}[c]{0.48\linewidth}
    \centering
    \includegraphics[width=\linewidth]{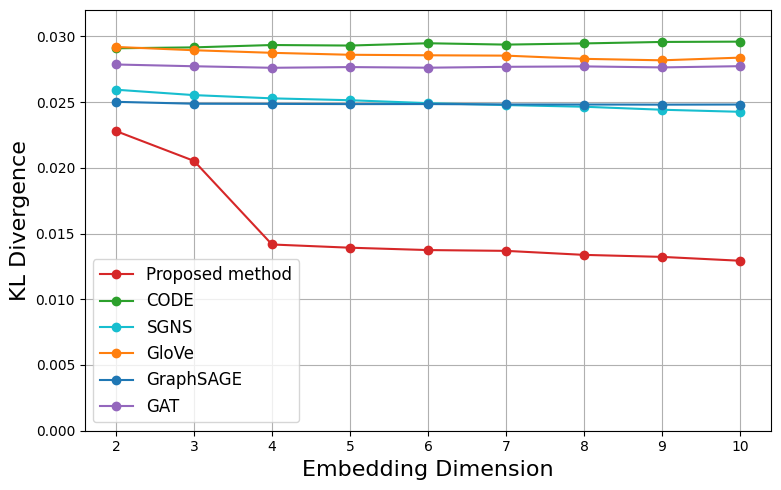}\\
    (b)
  \end{minipage} 
  \caption{Quantitative evaluation using KL divergence (Lower is better). (a) Adjective--noun data; (b) NeurIPS document--term data. }
  \figlabel{KL-evaluation}
\end{figure}

\begin{figure}[tp]
  \centering
  \includegraphics[width=\linewidth]{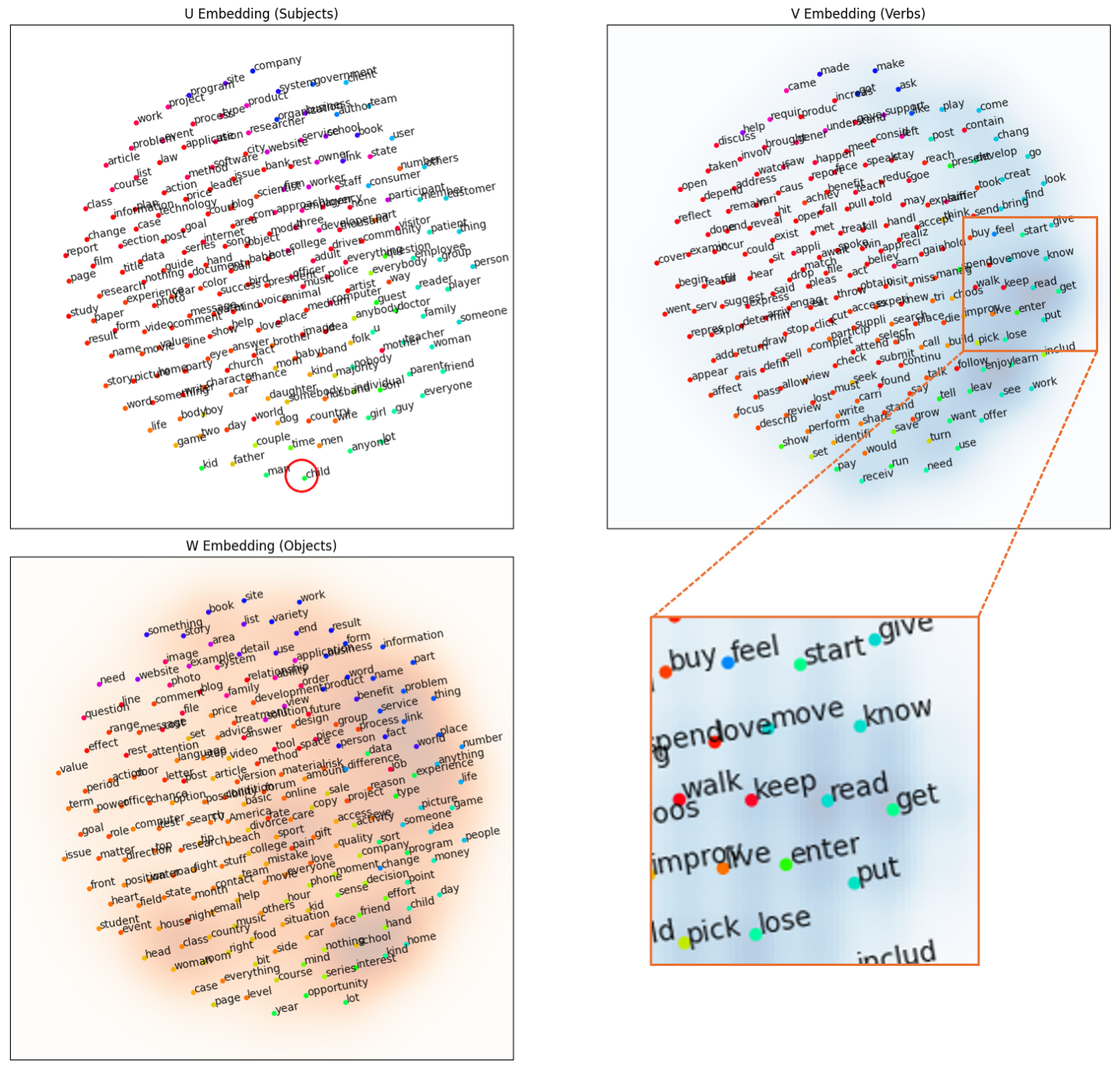}
  \caption{Visualization results on the subject–verb–object co-occurrence data.  
Top left: subject map; top right: verb map; bottom left: object map.  
The heatmap shows conditional co-occurrence probabilities using Coloring-by-Conditional-Probability (CbCP), with the target of interest set to ``\texttt{subject = child}.''}
  \figlabel{3domain-a}
\end{figure}
\begin{figure}[tp]
  \centering
  \includegraphics[width=\linewidth]{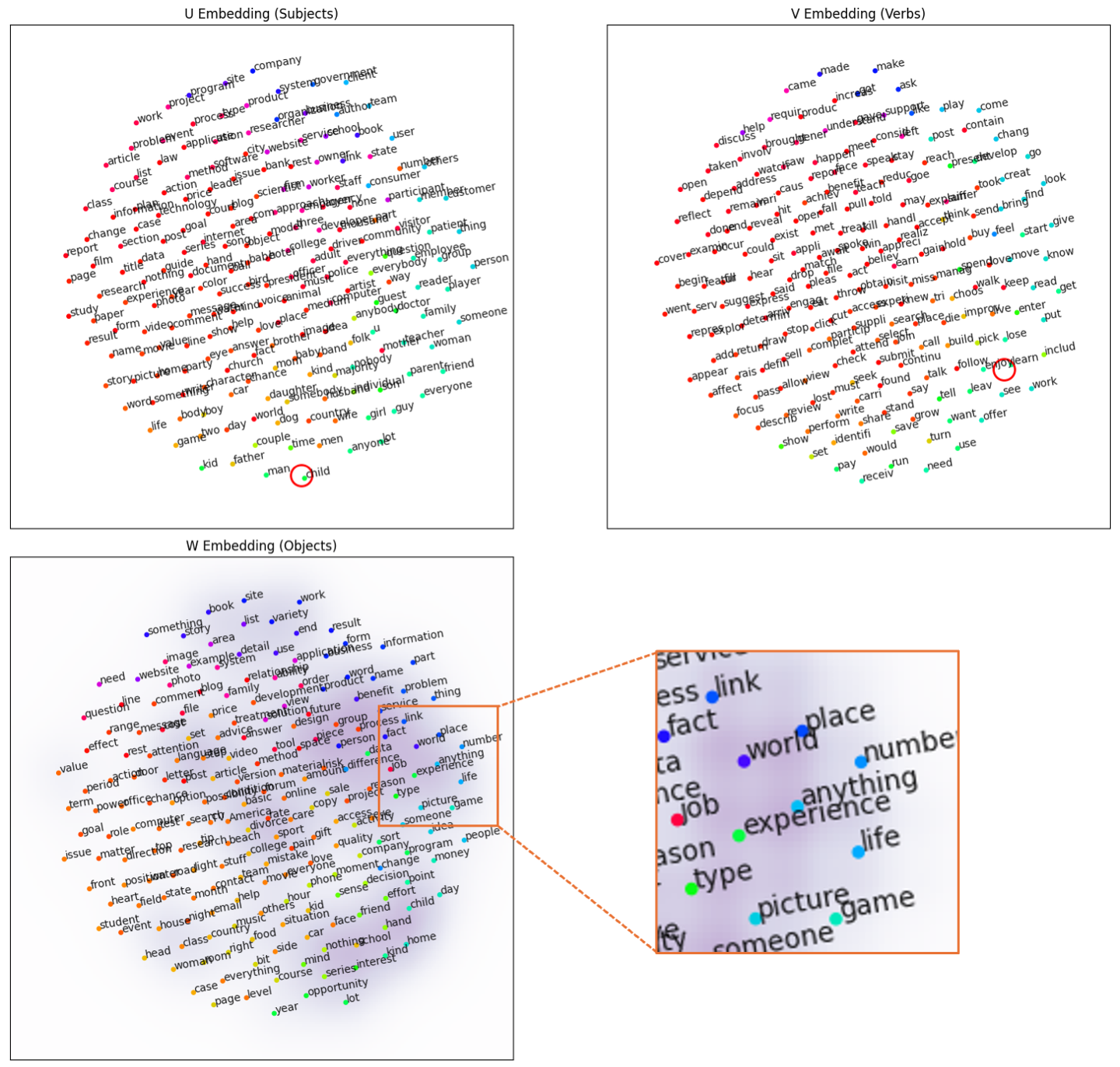}
\caption{
Visualization results on the subject–verb–object co-occurrence data.  
Top left: subject map; top right: verb map; bottom left: object map.  
The heatmap shows conditional co-occurrence probabilities using Coloring-by-Conditional-Probability (CbCP), with the targets of interest set to ``\texttt{subject = child, verb = learn}.''
}
  \figlabel{3domain-b}
\end{figure}

\subsection{Quantitative evaluation}

To assess the accuracy of the proposed method in modeling co-occurrence probabilities, we conducted a quantitative comparison with several representative embedding methods, including both co-occurrence-based and graph-based approaches. Since our primary goal is to support visual exploration, we focused particularly on extremely low-dimensional embeddings, especially in two dimensions. We evaluated the quality of each method using the KL divergence between the empirical co-occurrence distribution $P(c \mid a_i, b_j)$ and the modeled distribution $Q(c \mid a_i, b_j)$. Details of the experimental settings are provided in Appendix~B.

\figref{KL-evaluation} shows the KL divergence of the proposed method across embedding dimensions from 2 to 10. To provide an appropriate model granularity for density estimation, we employed an adaptive kernel bandwidth based on the rule-of-thumb estimator \cite{Scott20151}, with a lower bound determined by a minimum number of effective neighbors ($n_{\text{min}}=3$) to avoid overfitting in high-dimensional settings. 

As the embedding dimension increases, the KL divergence decreases notably until $d=4$, where it reaches its minimum. This optimal dimension $d=4$ likely reflects the limit of density estimation accuracy given the sample size (200--500). Beyond this point, the performance plateaus or slightly degrades, indicating over-smoothing due to the curse of dimensionality in kernel density estimation. 

We compared our method with several baselines: CODE (a heterogeneous co-occurrence embedding), Skip-Gram with Negative Sampling (SGNS, word2vec), GloVe (a word2vec-style model), GraphSAGE (a deep graph embedding), and GAT (a graph attention network). Across both evaluation metrics, our method consistently outperformed all baselines. Performance differences among the baseline methods were minimal, and their scores remained largely unchanged across embedding dimensions from 2 to 10.

It is important to note, however, that this comparison does not reflect the original design objectives of these baseline methods. Most of them were developed for downstream tasks such as classification, and not for directly modeling co-occurrence probabilities. Furthermore, they generally assume higher-dimensional embeddings (e.g., 100+ dimensions), and are not optimized for the extremely low-dimensional settings emphasized in our evaluation. Therefore, the lack of performance variation should not be interpreted as an inherent limitation of the methods themselves, but rather as a mismatch between their intended usage and the present evaluation setting.

In summary, the results demonstrate that the proposed method provides a theoretically grounded approximation of co-occurrence distributions when sufficient data is available. It performs particularly well in extremely low-dimensional settings (2--3D), where conventional methods tend to struggle. This makes it especially suitable for applications that require interpretable visualizations and interactive exploration of heterogeneous co-occurrence data, while conventional methods remain advantageous in high-dimensional settings for predictive tasks.

\subsection{Three-domain co-occurrence data}

The proposed method naturally extends beyond two-domain co-occurrence. Since it is formulated based on a generalized information-theoretic objective, it can accommodate higher-order structures without requiring any modification to the core algorithm. In this experiment, we applied the method to three-domain co-occurrence data consisting of subject--verb--object (SVO) triplets.

We used SVO data extracted from the NELL project,\footnote{\url{http://rtw.ml.cmu.edu/resources/svo/}} selecting 200 entries from each domain to construct a $200\times 200 \times 200$ co-occurrence tensor. Due to the sparsity of triadic co-occurrence, we applied Markov diffusion smoothing to estimate indirect associations. \figref{3domain-a} and \figref{3domain-b} present the resulting embeddings and visualizations.

In the subject map, the lower-right region includes human-related subjects such as \emph{man}, \emph{woman}, and \emph{child}, while the upper region contains organizational entities such as \emph{company}, \emph{organization}, and \emph{government}. On the left side, document-related nouns such as \emph{report}, \emph{page}, \emph{title}, and \emph{section} are located. Similarly, in the verb map, the right side contains verbs typically associated with human subjects, including \emph{enjoy}, \emph{learn}, \emph{look}, and \emph{know}, while the left side includes more abstract or impersonal verbs such as \emph{cover}, \emph{depend}, and \emph{reflect}.

We also applied the CbCP method to visualize conditional relationships among domains. In Figure~\ref{fig:3domain-a}, the subject \emph{child} is selected as the ToI, and the conditional probabilities are displayed over the verb and object maps. The highlighted verbs include \emph{keep}, \emph{know}, \emph{enter}, and \emph{walk}, all of which plausibly align with human agents. On the object map, terms such as \emph{name}, \emph{word}, and \emph{world} appear as frequently co-occurring objects, suggesting a context of learning or interaction.

In Figure~\ref{fig:3domain-b}, we fix both the subject (\emph{child}) and verb (\emph{learn}) to visualize the conditional distribution over the object domain. The highlighted region includes words such as \emph{experience}, \emph{world}, \emph{place}, and \emph{number}, which collectively indicate a plausible semantic field related to child learning. These results illustrate that our method enables the exploration of multi-way associations and latent context structures, even in sparse three-domain settings. Such role-specific interactions across three domains cannot be adequately represented by embedding all elements into a single latent space or by relying solely on pairwise similarity measures.

\section{Discussion}

\subsection{Theoretical Positioning and Applicability}

Our method provides a principled framework for modeling heterogeneous co-occurrence structures in low-dimensional latent spaces. Central to this framework are Lemmas~1 and~2, which derive the optimal co-occurrence density $q(u, v)$ in closed form as a mixture over kernel functions centered at latent coordinates. This kernel-based formulation is not an arbitrary choice, but a consequence of minimizing the KL divergence under a variational approximation. Thus, kernel density estimation emerges naturally as the theoretically optimal solution within our setting.

Importantly, our method is tailored for extremely low-dimensional embeddings (e.g., 2D or 3D), where interpretability and human-in-the-loop analysis are essential. This objective fundamentally differs from that of dot-product-based methods such as word2vec or matrix factorization, which prioritize preserving linear structure and often rely on high-dimensional spaces to be effective. These two aims---nonlinear compression into interpretable low dimensions versus linear algebraic structure in high dimensions---are not merely different, but theoretically incompatible. If co-occurrence information is to be maximally compressed into a compact nonlinear space, it is natural and even inevitable that linear properties such as vector arithmetic or similarity under inner products will no longer hold.

In this sense, our method prioritizes nonlinear structural fidelity over linear expressivity. It serves as a complementary tool to expressive models like GraphSAGE or GAT, which excel in learning task-oriented, high-dimensional representations. While such models aim for predictive performance, our approach emphasizes theoretical clarity and structural insight. Future work may explore hybrid architectures where high-capacity models feed into our framework for visualization---though this may forgo some of the closed-form guarantees central to our method.

\subsection{Limitations and Weaknesses}

While theoretically grounded, our method inherits practical limitations associated with kernel density estimation. Chief among them is the \emph{curse of dimensionality}; in higher-dimensional latent spaces, density estimation becomes statistically fragile due to data sparsity. Although the computational cost scales linearly with dimension, reliable estimation requires an increasing number of support points, which limits the scalability and robustness of the model in sparse regions.

Another inherent limitation is the \emph{lack of linear compositionality} in the learned embeddings. In contrast to inner-product-based models, our embeddings are not directly suited for linear similarity computation or vector arithmetic. However, this reflects a deliberate design decision rather than a deficiency. The method is specifically intended to construct dense, nonlinear, and interpretable low-dimensional representations. As shown in Lemma~1, the resulting density minimizes a variational upper bound of the KL divergence, thereby offering a theoretically justified structure.

In summary, while our method is not applicable to tasks that require high-dimensional expressivity or algebraic manipulation, it fills a distinctive role in interpretable, low-dimensional modeling. This is particularly beneficial for tasks that involve human-centered analysis, exploratory search, or the interpretation of asymmetric co-occurrence patterns.

\subsection{Visual Exploration Guided by Asymmetric Co-occurrence Modeling}

A central aim of our method is to support exploratory analysis through low-dimensional embeddings that are not only interpretable but also navigable. Conventional visualization techniques such as multidimensional scaling (MDS) or t-SNE typically focus on preserving local distances or neighborhood relationships in a static two-dimensional layout. While these methods are effective in revealing clustered structures in symmetric similarity data, they offer limited support for interactive exploration or directed inference, as they lack an explicit probabilistic model.

In contrast, our method enables a fundamentally different mode of interaction: users can trace semantic transitions by following the co-occurrence probability structure modeled explicitly in the latent space. Since our model captures the conditional probability $q(c \mid u,v)$, it enables a Markov-like exploration process, in which the analyst can move from one node to another based on probabilistic associations. This supports what we call \emph{semantic teleportation}---navigating across distant regions of the map via meaningful co-occurrence links, rather than relying solely on geometric proximity. Such a navigation paradigm is not feasible in conventional methods like t-SNE or CODE, which do not model co-occurrence probabilities explicitly and therefore cannot support directed transitions.

This behavior is particularly effective for asymmetric co-occurrence data, where a single concept may relate to multiple, qualitatively different clusters. In such cases, the exploration naturally branches, reflecting multiple semantic neighborhoods. That is, \emph{asymmetry introduces a branching structure}, which allows the Markov walk to diverge and guide the analyst toward unexpected yet contextually relevant regions—potentially enabling a form of \emph{serendipitous discovery}.

Moreover, prior work has suggested that higher-order semantic signals are often embedded in the \emph{asymmetric} components of co-occurrence patterns~\cite{Geffet2005107,Levy2014302}. Because our model maintains separate latent spaces for each domain, it preserves these directional cues by design. In contrast, approaches based on a shared embedding and symmetric similarity—such as dot-product models—tend to obscure such asymmetric or higher-order relationships. Our framework instead enables the analysis of \emph{directed semantic flows}, enhancing interpretability and expanding the range of inferences that can be drawn through interactive visualization.

To the best of our knowledge, this is the first visual analytics framework that combines an explicitly asymmetric co-occurrence model with guided exploration via Markov transitions in a low-dimensional embedding. This fills a methodological gap identified in prior literature (see Section~\ref{sec:asymmetry}) and opens up new opportunities for exploratory data analysis, particularly in heterogeneous or conceptually diffuse domains.

\subsection{Computational Cost and Scalability}

Co-occurrence data is combinatorial in nature, and the data size grows with the product of domain sample sizes: for two-domain data, the number of entries is $N = N_A \times N_B$.  
However, the computational cost of our method is only $O(N_A^2 + N_B^2)$,  
as kernel evaluations are performed independently within each domain.  
This cost remains affordable even on standard CPUs.

To evaluate scalability, we conducted an experiment using a three-domain SVO dataset  
with $200 \times 200 \times 200 = 8$ million possible triplets.  
Although the number of entries is large, the computation involves only  
$O(N_S^2 + N_V^2 + N_O^2)$ operations.  
The full embedding (200 iterations) was completed in under 8 minutes on a standard desktop machine (Mac mini with M4 chip), without GPU acceleration.  
This demonstrates that our method is lightweight and practical for moderate-scale data.

Scalability can be further improved using Nystr\"om-type kernel approximation~\cite{Miyazaki2022},  
which reduces the cost from $O(N_A^2 + N_B^2)$ to $O(N_A m + N_B m)$, where $m \ll N_A, N_B$.  
This technique is compatible with our framework and can be integrated without changing its theoretical foundation.

\subsection{Generalization to Multi-Domain and Hypergraph Embedding}

Higher-order co-occurrence structures, such as subject–verb–object (SVO) triplets, cannot be adequately captured by conventional embedding approaches based on pairwise similarity or inner products in a shared latent space. Unlike bipartite co-occurrence, typically modeled by matrix factorization or shared embeddings, triadic structures involve inherently asymmetric, role-specific relationships. These are more naturally represented as hyperedges in a hypergraph, where each domain (e.g., subject, verb, object) plays a distinct semantic role that should not be collapsed into a unified representation.

Prior studies have addressed SVO co-occurrence using tensor decompositions \cite{VanDeCruys2010417,VanDeCruys20122703} or knowledge graph embeddings \cite{Nickel2011809,Balazevic20195185}. While some of these approaches assign distinct embeddings to each domain, they often operate within a shared latent space and rely on bilinear or trilinear forms. This tends to obscure role-specific semantics and limits interpretability. Recent hypergraph embedding methods \cite{Feng20193558,Bai2021} support higher-order relations; however, they typically assume homogeneous node types and thus do not preserve domain asymmetry or enable conditional visual exploration.

In contrast, our method maintains separate latent spaces for each domain and estimates asymmetric joint and conditional distributions over multi-domain co-occurrence. This allows flexible, interpretable exploration of heterogeneous relationships via techniques such as Coloring-by-Conditional-Probability (CbCP). To the best of our knowledge, this is the first embedding approach that (i) preserves role-specific structure, (ii) avoids shared latent spaces, and (iii) supports conditional exploration in multi-domain settings. We view this as a step toward hypergraph-style co-occurrence modeling, where interpretability and asymmetry are first-class design principles.

\section{Conclusion}

In this paper, we proposed a novel method for embedding heterogeneous co-occurrence data into distinct latent spaces, enabling effective visual information exploration. Our approach models the co-occurrence probability density $q(u, v)$ or its conditional form $q(c \mid u, v)$, and embeds different domains---such as nouns and adjectives---into separate latent coordinate systems. This design allows for expressive and interpretable visualization of non-symmetric relationships, which are common in many real-world datasets.

From a theoretical standpoint, we showed that minimizing the upper bound of the KL divergence $D_{\mathrm{KL}}[P \| Q]$, where $Q(a_i,b_j)$ is the reconstructed co-occurrence probability from the embedding, leads naturally to a kernel density estimation framework. This theoretical grounding ensures that our embedding is not only visually intuitive but also statistically principled.

Through experiments on adjective--noun and word--paper co-occurrence datasets, we demonstrated the effectiveness of our method in reconstructing co-occurrence probabilities with low-dimensional embeddings. Furthermore, the ability to explore asymmetric conditional relationships using dual latent spaces supports novel forms of visual analysis based on conditional maps and Markov-like exploration.

We further applied our framework to three-way subject--verb--object (SVO) co-occurrence data, demonstrating its extensibility to multi-entity relational structures. By modeling higher-order conditional probabilities, our method enables the analysis of indirect or transitive associations and provides a basis for completing sparse co-occurrence patterns via Markov-based inference.

Future work includes extending our framework to handle multi-way co-occurrence data and incorporating side information or semantic constraints. We also plan to explore interactive visualization systems that take full advantage of the dual-space embedding for human-in-the-loop discovery.

\section*{Acknowledgments}
We are grateful to Dr. Ishibashi for his insightful advice and generous support throughout this study. This work was partially supported by ZOZO Research and JSPS Kakenhi, Grant number 21K12061.

\bibliographystyle{ijicicbib}
\bibliography{ref}

\begin{thebibliography}{10}

\bibitem{Dagan199943}
I.~Dagan, L.~Lee, and F.~Pereira, ``Similarity-based models of word cooccurrence probabilities,'' {\em Machine Learning}, \textbf{34} (1) (1999) 43--69.

\bibitem{Weeds2005439}
J.~Weeds and D.~Weir, ``Co-occurrence retrieval: A flexible framework for lexical distributional similarity,'' {\em Computational Linguistics}, \textbf{31} (4) (2005) 439--475.

\bibitem{Sanderson2009771}
J.~Sanderson, J.~Diamond, and S.~Pimm, ``Pairwise co-existence of bismarck and solomon landbird species,'' {\em Evolutionary Ecology Research}, \textbf{11} (5) (2009) 771--786.

\bibitem{Small1973265}
H.~Small, ``Co-citation in the scientific literature: A new measure of the relationship between two documents,'' {\em Journal of the American Society for Information Science}, \textbf{24} (4) (1973) 265--269.

\bibitem{Leydesdorff20061616}
L.~Leydesdorff and L.~Vaughan, ``Co-occurrence matrices and their applications in information science: Extending {ACA} to the web environment,'' {\em Journal of the American Society for Information Science and Technology}, \textbf{57} (12) (2006) 1616--1628.

\bibitem{Otte2002441}
E.~Otte and R.~Rousseau, ``Social network analysis: A powerful strategy, also for the information sciences,'' {\em Journal of Information Science}, \textbf{28} (6) (2002) 441--453.

\bibitem{Chen2012714}
S.~Chen, J.~Moore, D.~Turnbull, and T.~Joachims, ``Playlist prediction via metric embedding,'' {\em Proceedings of the ACM SIGKDD International Conference on Knowledge Discovery and Data Mining}.

\bibitem{Fu20071}
X.~Fu, X.~Shen, S.-H. Hong, Y.~Wu, N.~Nikolov, and K.~Xu, ``Visualization and analysis of email networks,'' {\em Asia-Pacific Symposium on Visualisation 2007, APVIS 2007, Proceedings}.

\bibitem{Wang2021}
H.~Wang, H.~Mai, Z.-H. Deng, C.~Yang, L.~Zhang, and H.-Y. Wang, ``Distributed representations of diseases based on co-occurrence relationship,'' {\em Expert Systems with Applications}, \textbf{183} (2021).

\bibitem{Mahecha200931}
M.~Mahecha, A.~Martínez, H.~Lange, M.~Reichstein, and E.~Beck, ``Identification of characteristic plant co-occurrences in neotropical secondary montane forests,'' {\em Journal of Plant Ecology}, \textbf{2} (1) (2009) 31--41.

\bibitem{Kohara201347}
Y.~Kohara and K.~Yanai, ``Visual analysis of tag co-occurrence on nouns and adjectives,'' {\em Lecture Notes in Computer Science (including subseries Lecture Notes in Artificial Intelligence and Lecture Notes in Bioinformatics)}, \textbf{7732 LNCS} (PART 1) (2013) 47--57.

\bibitem{Zhang2016}
H.~Zhang, X.~Shang, H.~Luan, M.~Wang, and T.-S. Chua, ``Learning from collective intelligence: Feature learning using social images and tags,'' {\em ACM Transactions on Multimedia Computing, Communications and Applications}, \textbf{13} (1) (2016).

\bibitem{Liu2015867}
C.-H. Liu, Y.-L. Lin, W.-F. Cheng, and W.~Hsu, ``Exploiting word and visual word co-occurrence for sketch-based clipart image retrieval,'' {\em MM 2015 - Proceedings of the 2015 ACM Multimedia Conference}.

\bibitem{Globerson20072265}
A.~Globerson, G.~Chechik, F.~Pereira, and N.~Tishby, ``Euclidean embedding of co-occurrence data,'' {\em Journal of Machine Learning Research}, \textbf{8} (2007) 2265--2295.

\bibitem{Fang2016122}
H.~Fang, F.~Wu, Z.~Zhao, X.~Duan, Y.~Zhuang, and M.~Ester, ``Community-based question answering via heterogeneous social network learning,'' {\em 30th AAAI Conference on Artificial Intelligence, AAAI 2016}.

\bibitem{Liu20161774}
L.~Liu, W.~Cheung, X.~Li, and L.~Liao, ``Aligning users across social networks using network embedding,'' {\em IJCAI International Joint Conference on Artificial Intelligence}, \textbf{2016-January} (2016) 1774--1780.

\bibitem{Sarkar2007420}
P.~Sarkar, S.~Siddiqi, and G.~Gordon, ``A latent space approach to dynamic embedding of co-occurrence data,'' {\em Journal of Machine Learning Research}, \textbf{2} (2007) 420--427.

\bibitem{Maron2010nips}
Y.~Maron, M.~Lamar, and E.~Bienenstock, ``Sphere embedding: An application to part-of-speech induction,'' in {\em Advances in Neural Information Processing Systems 23: 24th Annual Conference on Neural Information Processing Systems 2010, NIPS 2010}, 2010.

\bibitem{Khoshneshin201087}
M.~Khoshneshin and W.~Street, ``Collaborative filtering via euclidean embedding,'' {\em RecSys'10 - Proceedings of the 4th ACM Conference on Recommender Systems}.

\bibitem{Khoshneshin201174}
M.~Khoshneshin, W.~N. Street, and P.~Srinivasan, ``Bayesian embedding of co-occurrence data for query-based visualization,'' in {\em Proceedings - 10th International Conference on Machine Learning and Applications, ICMLA 2011}, vol.~1, p.~74 – 79, 2011.

\bibitem{Xie2021}
Y.~Xie, B.~Yu, S.~Lv, C.~Zhang, G.~Wang, and M.~Gong, ``A survey on heterogeneous network representation learning,'' {\em Pattern Recognition}, \textbf{116} (2021).

\bibitem{White1981163}
H.~White and B.~Griffith, ``Author cocitation: A literature measure of intellectual structure,'' {\em Journal of the American Society for Information Science}, \textbf{32} (3) (1981) 163--171.

\bibitem{Levy20142177}
O.~Levy and Y.~Goldberg, ``Neural word embedding as implicit matrix factorization,'' {\em Advances in Neural Information Processing Systems}, \textbf{3} (January) (2014) 2177--2185.

\bibitem{Pennington20141532}
J.~Pennington, R.~Socher, and C.~Manning, ``Glove: Global vectors for word representation,'' {\em EMNLP 2014 - 2014 Conference on Empirical Methods in Natural Language Processing, Proceedings of the Conference}.

\bibitem{VanDeCruys2010417}
T.~Van De~Cruys, ``A non-negative tensor factorization model for selectional preference induction,'' {\em Natural Language Engineering}, \textbf{16} (4) (2010) 417--437.

\bibitem{Nickel2011809}
M.~Nickel, V.~Tresp, and H.-P. Kriegel, ``A three-way model for collective learning on multi-relational data,'' Proceedings of the 28th International Conference on Machine Learning, ICML 2011, p.~809 – 816, 2011.

\bibitem{Mikolov2013}
T.~Mikolov, K.~Chen, G.~Corrado, and J.~Dean, ``Efficient estimation of word representations in vector space,'' 1st International Conference on Learning Representations, ICLR 2013 - Workshop Track Proceedings, 2013.

\bibitem{Tang20151067}
J.~Tang, M.~Qu, M.~Wang, M.~Zhang, J.~Yan, and Q.~Mei, ``Line: Large-scale information network embedding,'' in {\em WWW 2015 - Proceedings of the 24th International Conference on World Wide Web}, p.~1067 – 1077, 2015.

\bibitem{Grover2016855}
A.~Grover and J.~Leskovec, ``Node2vec: Scalable feature learning for networks,'' in {\em Proceedings of the ACM SIGKDD International Conference on Knowledge Discovery and Data Mining}, vol.~13-17-August-2016, p.~855 – 864, 2016.

\bibitem{Hamilton20171025}
W.~L. Hamilton, R.~Ying, and J.~Leskovec, ``Inductive representation learning on large graphs,'' in {\em Advances in Neural Information Processing Systems}, vol.~2017-December, p.~1025 – 1035, 2017.

\bibitem{Velickovic2018}
P.~Veli\v{c}kovi\'c, A.~Casanova, P.~Liò, G.~Cucurull, A.~Romero, and Y.~Bengio, ``Graph attention networks,'' in {\em 6th International Conference on Learning Representations, ICLR 2018 - Conference Track Proceedings}, 2018.

\bibitem{Gries2013137}
S.~T. Gries, ``50-something years of work on collocations: What is or should be next \dots,'' {\em International Journal of Corpus Linguistics}, \textbf{18} (1) (2013) 137 – 166.

\bibitem{Levy2014302}
O.~Levy and Y.~Goldberg, ``Dependency-based word embeddings,'' 52nd Annual Meeting of the Association for Computational Linguistics, ACL 2014 - Proceedings of the Conference, vol.~2, p.~302 – 308, 2014.

\bibitem{Roller20141025}
S.~Roller, K.~Erk, and G.~Boleda, ``Inclusive yet selective: Supervised distributional hypernymy detection,'' in {\em COLING 2014 - 25th International Conference on Computational Linguistics, Proceedings of COLING 2014: Technical Papers}, p.~1025 – 1036, 2014.

\bibitem{Tissier2017254}
J.~Tissier, C.~Gravier, and A.~Habrard, ``Dict2vec : Learning word embeddings using lexical dictionaries,'' EMNLP 2017 - Conference on Empirical Methods in Natural Language Processing, Proceedings, p.~254 – 263, 2017.

\bibitem{Liang201659}
D.~Liang, J.~Altosaar, L.~Charlin, and D.~Blei, ``Factorization meets the item embedding: Regularizing matrix factorization with item co-occurrence,'' {\em RecSys 2016 - Proceedings of the 10th ACM Conference on Recommender Systems}.

\bibitem{Elkan2008213}
C.~Elkan and K.~Noto, ``Learning classifiers from only positive and unlabeled data,'' Proceedings of the ACM SIGKDD International Conference on Knowledge Discovery and Data Mining, p.~213 – 220, 2008.

\bibitem{Bekker2020719}
J.~Bekker and J.~Davis, ``Learning from positive and unlabeled data: a survey,'' {\em Machine Learning}, \textbf{109} (4) (2020) 719 – 760.

\bibitem{grewal20202196}
K.~Grewal and Y.~Xu, ``Chaining and historical adjective extension,'' in {\em Proceedings for the 42nd Annual Meeting of the Cognitive Science Society: Developing a Mind: Learning in Humans, Animals, and Machines, CogSci 2020}, p.~2196 – 2202, 2020.

\bibitem{Scott20151}
D.~W. Scott, {\em Multivariate density estimation: Theory, practice, and visualization: Second edition}.
\newblock Wiley, 1992.

\bibitem{Geffet2005107}
M.~Geffet and I.~Dagan, ``The distributional inclusion hypotheses and lexical entailment,'' in {\em ACL-05 - 43rd Annual Meeting of the Association for Computational Linguistics, Proceedings of the Conference}, p.~107 – 114, 2005.

\bibitem{Miyazaki2022}
K.~Miyazaki, S.~Takano, R.~Tsuno, H.~Ishibashi, and T.~Furukawa, ``Low-rank kernel decomposition for scalable manifold modeling,'' in {\em 2022 Joint 12th International Conference on Soft Computing and Intelligent Systems and 23rd International Symposium on Advanced Intelligent Systems, SCIS and ISIS 2022}, 2022.

\bibitem{VanDeCruys20122703}
T.~Van De~Cruys, L.~Rimell, T.~Poibeau, and A.~Korhonen, ``Multi-way tensor factorization for unsupervised lexical acquisition,'' in {\em 24th International Conference on Computational Linguistics - Proceedings of COLING 2012: Technical Papers}, p.~2703 – 2720, 2012.

\bibitem{Balazevic20195185}
I.~Bala\v{z}evi\'c, C.~Allen, and T.~M. Hospedales, ``Tucker: Tensor factorization for knowledge graph completion,'' in {\em EMNLP-IJCNLP 2019 - 2019 Conference on Empirical Methods in Natural Language Processing and 9th International Joint Conference on Natural Language Processing, Proceedings of the Conference}, p.~5185 – 5194, 2019.

\bibitem{Feng20193558}
Y.~Feng, H.~You, Z.~Zhang, R.~Ji, and Y.~Gao, ``Hypergraph neural networks,'' in {\em 33rd AAAI Conference on Artificial Intelligence, AAAI 2019, 31st Innovative Applications of Artificial Intelligence Conference, IAAI 2019 and the 9th AAAI Symposium on Educational Advances in Artificial Intelligence, EAAI 2019}, p.~3558 – 3565, 2019.

\bibitem{Bai2021}
S.~Bai, F.~Zhang, and P.~H. Torr, ``Hypergraph convolution and hypergraph attention,'' {\em Pattern Recognition}, \textbf{110} (2021).

\end{thebibliography}

\section*{Appendix A. Supplementary Proofs for Lemmas}

\subsection*{A-1. Proof of Lemma 1}

This section provides a detailed derivation of Lemma~1 (Eq.~\eqref{lemma1}),  
which establishes an upper bound on the KL divergence \( \mathrm{D_{KL}}[P \| Q] \)  
between the empirical co-occurrence probability \( P_{ij} \) and the model-estimated probability \( Q_{ij} \),  
via the application of Jensen’s inequality.

\subsubsection*{\bf Preliminaries}

The proof involves both discrete item pairs \( (a_i, b_j) \) and their associated continuous latent representations \( (u, v) \in \mathcal{U} \times \mathcal{V} \).  
Let \( P_{ij} \equiv P(a_i, b_j) \) denote the empirical co-occurrence probability of the pair \( (a_i, b_j) \),  
with marginal probabilities \( P_i \equiv P(a_i) \) and \( P_j \equiv P(b_j) \).  
Each item \( a_i \) and \( b_j \) is embedded into the latent space via vectors \( u_i \in \mathcal{U} \) and \( v_j \in \mathcal{V} \), respectively.

We assume that the conditional distributions from items to the latent space are modeled by isotropic Gaussian kernels:
\[
p(u \mid a_i) = k(u \mid u_i) \equiv \mathcal{N}(u \mid u_i, \sigma^2 I), \quad
p(v \mid b_j) = k(v \mid v_j) \equiv \mathcal{N}(v \mid v_j, \sigma^2 I).
\]
By Bayes’ theorem, the reverse conditionals are given by:
\[
P(a_i \mid u) = \frac{k(u \mid u_i) \cdot P_i}{p(u)}, \quad
P(b_j \mid v) = \frac{k(v \mid v_j) \cdot P_j}{p(v)},
\]
where \( p(u) = \sum_i k(u \mid u_i)\, P_i \) and \( p(v) = \sum_j k(v \mid v_j)\, P_j \).

Substituting these expressions into the model definition, the model-estimated co-occurrence probability \( Q_{ij} \equiv Q(a_i, b_j) \) can be written as:
\begin{align}
Q_{ij}
&= \iint P(a_i \mid u)\, P(b_j \mid v)\, q(u, v)\, du\, dv \nonumber \\
&= P_i P_j \iint k(u \mid u_i)\, k(v \mid v_j)\, \frac{q(u, v)}{p(u)\, p(v)}\, du\, dv.
\eqlabel{A1}
\end{align}

To relate the model density \( q(u,v) \) to the empirical data,  
we define the following kernel density estimate over the latent space:
\[
p(u, v) \equiv \sum_{i,j} P_{ij}\, k(u \mid u_i)\, k(v \mid v_j).
\]
This density \( p(u, v) \) serves as a continuous relaxation of the discrete empirical distribution \( P_{ij} \),  
achieved via Gaussian smoothing in the latent space.  
The marginal distributions \( p(u) \) and \( p(v) \) are then naturally induced by integrating \( p(u, v) \) along each axis.

The goal of the proposed method is to learn a model distribution \( q(u, v) \)  
that approximates \( p(u, v) \) while preserving the structural properties of co-occurrence.

\subsubsection*{\bf KL divergence and its upper bound}

We now derive an upper bound on the KL divergence \( \mathrm{D_{KL}}[P \| Q] \)  
between the empirical distribution \( P_{ij} \) and the model-estimated distribution \( Q_{ij} \).  
Based on the formulation in Eq.~\eqref{A1}, we have:
\begin{align*}
\mathrm{D_{KL}}[P \| Q]
&= \sum_{i,j} P_{ij} \log \frac{P_{ij}}{Q_{ij}} \\
&= \sum_{i,j} P_{ij} \left( \log \frac{P_{ij}}{P_i P_j} - \log \frac{Q_{ij}}{P_i P_j} \right) \\
&= \sum_{i,j} P_{ij} \log \frac{P_{ij}}{P_i P_j}
  - \sum_{i,j} P_{ij} \log \left( \iint k(u \mid u_i)\, k(v \mid v_j)\, \frac{q(u,v)}{p(u)\,p(v)}\, du\, dv \right).
\end{align*}
The first term corresponds to the mutual information between \( a \) and \( b \) under the empirical distribution \( P \),  
denoted as \( I_P[a; b] \).
To upper-bound the second term, we apply Jensen’s inequality.  
Let us define the weight function as the product of Gaussian kernels:
\[
w(u, v) = k(u \mid u_i)\, k(v \mid v_j),
\]
and the integrand function as:
\[
f(u, v) = \frac{q(u,v)}{p(u)\,p(v)}.
\]
Since \( \log(\cdot) \) is concave, Jensen’s inequality gives:
\[
- \log \left( \iint w(u, v)\, f(u, v)\, du\, dv \right)
\leq - \iint w(u, v)\, \log f(u, v)\, du\, dv.
\]
Applying this inequality to each pair \( (i, j) \) and summing over \( P_{ij} \), we obtain:
\begin{align*}
&- \sum_{i,j} P_{ij} \log \left( \iint k(u \mid u_i)\, k(v \mid v_j)\, \frac{q(u,v)}{p(u)\,p(v)}\, du\, dv \right) \\
&\qquad \leq - \sum_{i,j} P_{ij} \iint k(u \mid u_i)\, k(v \mid v_j)\, \log \frac{q(u,v)}{p(u)\,p(v)}\, du\, dv.
\end{align*}
Thus, we obtain the following upper bound:
\begin{align*}
\mathrm{D_{KL}}[P \| Q]
&\leq I_P[a; b] - \sum_{i,j} P_{ij} \iint k(u \mid u_i)\, k(v \mid v_j)\, \log \frac{q(u,v)}{p(u)\,p(v)}\, du\, dv.
\end{align*}
Now, since \( p(u,v) \) is defined as the kernel density estimate in Eq.~\eqref{eq:kernel_density},  
we can rewrite the second term as an expectation under \( p(u,v) \):
\[
\sum_{i,j} P_{ij} \iint k(u \mid u_i)\, k(v \mid v_j)\, \log \frac{q(u,v)}{p(u)\,p(v)}\, du\, dv
= \iint p(u,v)\, \log \frac{q(u,v)}{p(u)\,p(v)}\, du\, dv.
\]
Substituting this into the upper bound yields:
\[
\mathrm{D_{KL}}[P \| Q]
\leq I_P[a; b] - \iint p(u,v)\, \log \frac{q(u,v)}{p(u)\,p(v)}\, du\, dv.
\]
We now rewrite the integrand by inserting \( p(u,v) \) into the ratio:
\begin{align*}
& I_P[a; b] - \iint p(u,v)\, \log \frac{q(u,v)}{p(u)\,p(v)}\, du\, dv \\
&= I_P[a; b] - \iint p(u,v)\, \log \left( \frac{q(u,v)}{p(u,v)} \cdot \frac{p(u,v)}{p(u)\,p(v)} \right)\, du\, dv \\
&= I_P[a; b] - \iint p(u,v)\, \log \frac{p(u,v)}{p(u)\,p(v)}\, du\, dv
    + \iint p(u,v)\, \log \frac{p(u,v)}{q(u,v)}\, du\, dv \\
&= I_P[a; b] - I_p[u; v] + \mathrm{D_{KL}}[p \| q].
\end{align*}
This completes the proof of Lemma~1.  
\qed

\vspace{1em}
\subsection*{A-2. Proof of Lemma 3}

Lemma~3 is a natural extension of Lemma~1 to the three-domain setting.  
The proof follows the same structure, with mutual information replaced by total correlation.

Let \( P_{ijk} = P(a_i, b_j, c_k) \) denote the empirical co-occurrence distribution,  
with marginals \( P_i = P(a_i) \), \( P_j = P(b_j) \), \( P_k = P(c_k) \).  
The total correlation is defined as:
\[
I_P[a; b; c] = \sum_{i,j,k} P_{ijk} \log \frac{P_{ijk}}{P_i P_j P_k}.
\]

Using the same kernel formulation as in Eq.~\eqref{Qijk}, and applying Jensen’s inequality as in Lemma~1, we obtain the upper bound:
\[
\mathrm{D_{KL}}[P \| Q] \leq I_P[a; b; c] - I_p[u; v; w] + \mathrm{D_{KL}}[p \| q],
\]
where \( p(u,v,w) \) is the kernel density estimate based on \( P_{ijk} \), and \( I_p[u;v;w] \) is the total correlation in latent space.

\subsubsection*{\bf Special case: when the upper bound is tight}

As in the two-domain case, the upper bound becomes tight when the model density matches the empirical kernel density:
\[
q(u,v,w) = p(u,v,w) = \sum_{i,j,k} P_{ijk}\, k(u \mid u_i)\, k(v \mid v_j)\, k(w \mid w_k).
\]
In this case, we have \( \mathrm{D_{KL}}[p \| q] = 0 \) and \( I_q[u; v; w] = I_p[u; v; w] \),  
yielding the simplified bound:
\[
\mathrm{D_{KL}}[P \| Q] \leq I_P[a; b; c] - I_q[u; v; w].
\]
\qed

\vspace{1em}
\subsection*{A-3. Proof of Lemma 4}

Lemma~4 extends Lemma~1 by incorporating the binary co-occurrence variable \( c \in \{0,1\} \).  
We denote the empirical joint distribution as \( P(c, a_i, b_j) \), with marginals  
\( P_i = P(a_i) \), \( P_j = P(b_j) \), and \( P_c = P(c) \).  
The model distribution is defined as:
\[
Q(c, a_i, b_j) = P_i P_j \iint k(u \mid u_i)\, k(v \mid v_j)\, \frac{q(c, u, v)}{p(u)\,p(v)}\, du\, dv.
\]

Applying Jensen’s inequality as in Lemma~1, we obtain:
\[
\mathrm{D_{KL}}[P \| Q] \leq I_P[c; a; b] - I_p[c; u; v] + \mathrm{D_{KL}}[p \| q],
\]
where \( p(c, u, v) \) is the kernel-smoothed empirical distribution,  
and \( I_p[c; u; v] \) denotes the mutual information in latent space.

The bound becomes tight when the model matches the kernel density estimate:
\[
q(c, u, v) = p(c, u, v) = \sum_{i,j} P(c, a_i, b_j)\, k(u \mid u_i)\, k(v \mid v_j),
\]
in which case we have:
\[
\mathrm{D_{KL}}[P \| Q] \leq I_P[c; a; b] - I_q[c; u; v].
\]
\qed

\section*{Appendix B. Experimental Settings}
\subsection*{B-1. Common Setting}

\subsubsection*{\bf Estimating co-occurrence probability via PU Learning}
Except for the artificial dataset, we estimated the co-occurrence probabilities \(P(c \mid a_i, b_j)\) from observed co-occurrence counts \(N_1(a_i, b_j)\) using a positive-unlabeled (PU) learning approach \cite{Elkan2008213,Bekker2020719}.

Estimating \(P(c=1 \mid a_i, b_j)\) is equivalent to estimating the mean \(\mu_{ij}\) of a Bernoulli distribution \(\mathrm{Ber}(c \mid \mu_{ij})\).  
Assuming a Beta prior \(\mathrm{Beta}(\alpha{+}1, \beta{+}1)\), the MAP estimate becomes:
\[
P(c=1 \mid a_i, b_j) = \frac{N_1(a_i, b_j) + \alpha}{N_1(a_i, b_j) + N_0(a_i, b_j) + \alpha + \beta},
\]
where \(N_0(a_i, b_j)\) is the estimated number of negative (non-co-occurring) instances.

Since \(N_0(a_i, b_j)\) is not directly observable, we approximate it under an independence assumption:
\[
N_0(a_i, b_j) = \frac{\beta}{\alpha} \cdot N_1 \cdot P(a_i \mid c{=}1) \cdot P(b_j \mid c{=}1),
\]
where \(P(a_i \mid c{=}1)\) and \(P(b_j \mid c{=}1)\) are the marginal frequencies estimated from the observed positive data, and $N_1=\sum_{i,j} N(a_i, b_j)$ is the total number of co-occurrence events.

This yields a Bayesian-smoothed estimate that is more stable than directly normalizing \(N_1(a_i, b_j)\).  
We typically set \(\alpha = 1\), \(\beta = 10\), and compute the complementary probability as:
\[
P(c=0 \mid a_i, b_j) = 1 - P(c=1 \mid a_i, b_j).
\]

\subsubsection*{\bf Regularization, optimization and initialization}
We applied L2 regularization \(\lambda \| u_i \|^2\) with \(\lambda = 0.01\), except for the artificial data. Optimization was performed using steepest gradient descent. When using random initialization, we adopted a noisy variant of SGD.

Latent vectors were initialized in two ways: by PCA or by sampling from a Gaussian distribution. In both cases, the vectors were placed near the origin, with their scale set between \(1/100\) and \(1/10\) of the kernel bandwidth.  
This setting allows embeddings to gradually diffuse outward during training, creating a simulated annealing effect, where initially small kernel influence helps avoid poor local optima in early stages.

\subsubsection*{\bf Parameter settings}
The kernel bandwidth \(\sigma\) was selected according to data density, typically set to about \(1/20\) of the final embedding variance. This ensured that each kernel neighborhood contained only a few samples, allowing the kernel to capture fine-grained local structures.

Following Section~3.3, we adopted progressive training: the auxiliary objective and the main objective were optimized sequentially for about 100 iterations each. These numbers were chosen conservatively and could be reduced depending on convergence in practice.

\subsection*{B-2. Experiment of Artificial Data}

We generated an artificial co-occurrence function \( f(u, v) \in [0, 1] \), which serves as the ground truth probability \( p(c = 1 \mid u, v) \).  
The latent variables were uniformly sampled from the interval \([0, 1]\).  
Initialization was done by sampling from a Gaussian distribution centered at the origin, and training was performed using noisy stochastic gradient descent.  
In this experiment, L2 regularization was not applied; instead, the embedding vectors were clipped to the range \([-1, +1]\).

\subsection*{B-3. Adjective--Noun Co-occurrence}

We used adjective–noun co-occurrence data derived from \cite{grewal20202196}.  
The adjective set consists of 200 items selected in the original dataset.  
For nouns, we selected 250 items based on their co-occurrence frequency with the fixed noun set, centered around the median.  
Adjectives with extremely low or extremely high co-occurrence frequencies were excluded to avoid noise and formulaic expressions.

For the visualizations shown in the main paper, initialization was performed using the first and second principal components obtained via PCA.  
While random initialization yielded similar overall tendencies, the PCA-based initialization was more stable and representative across trials.

\subsection*{\bf B-4. NeurIPS Paper–Term Co-occurrence}

We used the NeurIPS conference paper dataset from 2001 to 2003, available at \url{https://cs.nyu.edu/~roweis/data.html}.  
The original bag-of-words data contains 593 documents and 14,036 unique words.  
We selected 436 terms using two criteria: (1) terms with co-occurrence frequencies centered around the median, and (2) terms with high TF-IDF scores when treating each research area as a single document.  
Papers with negligible co-occurrence to the selected terms were excluded, resulting in a final set of 540 papers.

To compute \( P(c \mid \text{term}_i, \text{paper}_j) \) via positive-unlabeled learning, we assumed a uniform prior over papers, i.e., \( P(\text{paper}_j) = 1/N_\text{paper} \).  
Other training settings were the same as in the adjective--noun experiment.

\subsection*{B-5. Quantitative Evaluation}

We conducted quantitative evaluation using the -adjective--noun and NeurIPS paper--term datasets described in Sections B-3 and B-4.  
The goal was to compare the quality of embeddings in reconstructing co-occurrence probability distributions.

\subsubsection*{\bf Proposed method}

The proposed method was initialized using PCA as in the previous sections.  
The evaluation score of our method depends sensitively on the kernel bandwidth used in density estimation.  
In contrast, the learned embedding vectors are stable and largely insensitive to parameter choices.  
Starting from an over-smoothed initialization, the vectors gradually spread out and converge to a stable configuration.  
After that, further improvement in the evaluation score is mainly due to overfitting of the density model \( q(u,v) \), not actual changes in the embeddings.  
This distinction is critical: the KL divergence may keep decreasing even when the embeddings have already converged.

To fairly assess the embedding quality, we first fixed the embeddings and then selected kernel bandwidths for density estimation.  
We followed the rule-of-thumb bandwidth selection method \cite{Scott20151} as follows:
\begin{align*}
h_u &= \sigma_u \left(\frac{4}{d+2}\right)^{1/(d+4)} n_u^{-1/(d+4)}\\
h_v &= \sigma_v \left(\frac{4}{d+2}\right)^{1/(d+4)} n_v^{-1/(d+4)}.
\end{align*}
To prevent excessive overfitting in high dimensions, we also designed the bandwidth such that the average number of neighbors within the kernel radius remained above three.  
As a result, kernel bandwidths were adjusted adaptively depending on dimensionality.

We intentionally designed the evaluation to be conservative, in order to avoid overestimating our method's performance due to overfitting in the density estimation stage.

\paragraph{\bf Baseline methods}

We evaluated several baseline methods for comparison.  
In principle, if a baseline provides an estimate of co-occurrence frequency or the empirical distribution \(P(a_i,b_j)\), we used it directly.  
If the method yields embedding vectors, we defined the predicted co-occurrence probability as
\[
Q(c=1 \mid i,j) = \sigma(u_i^\top v_j),
\]
where \(\sigma\) is the sigmoid function.

If the inner products were poorly scaled or biased (e.g., SGNS embeddings are strictly non-negative due to truncation), we calibrated the output by fitting scaling and bias parameters \((a,b)\) using least squares:
\[
Q(c=1 \mid i,j) = \sigma(a \cdot (u_i^\top v_j - b)).
\]
For most methods, such calibration degraded performance, so we kept \((a,b) = (1,0)\) unless noted otherwise.  
Details and exceptions are described below.

\begin{description}
\item[CODE]
CODE estimates the conditional distribution \(Q(u,v \mid c=1)\) as a Gaussian:
\[
Q(u,v \mid c=1) \propto \exp\left(-\| u_i - v_j \|^2/2 \right).
\]
We converted this to \(Q(c=1 \mid u_i, v_j)\) using Bayes’ rule and normalization.

\item[SGNS (word2vec)]
We constructed the SPPMI matrix and applied truncated SVD to obtain embeddings.  
Since the inner product \(u_i^\top v_j\) is strictly non-negative due to truncation, the sigmoid output is inherently biased toward 1.  
To correct this, we fitted the parameters \((a,b)\) by least squares.  
This calibration was essential for fair comparison.

\item[GloVe]
We trained embeddings directly from the co-occurrence frequency data \(N_1(a_i, b_j)\) using the GloVe algorithm.  
This method estimates \(\log N_1(a_i, b_j)\) using embedding vectors and bias terms as \(u_i^\top v_j + b_i + b_j\).  
We trained the embeddings from raw frequency data and then used PU learning to estimate \(Q(c \mid a_i, b_j)\) from the predicted values.  
The inclusion of bias terms improved performance in the adjective--noun dataset, but not in the NeurIPS dataset; we adopted the best-performing setup in each case.

\item[GraphSAGE]
We constructed a bipartite graph from the co-occurrence frequency and used GraphSAGE from the PyTorch Geometric (PyG) library to learn embeddings.  
We then defined \(Q(c=1 \mid i,j) = \sigma(u_i^\top v_j)\).  
Attempting to fit \((a,b)\) degraded the performance due to the skewed distribution of inner products; we therefore kept \((a,b) = (1,0)\).

\item[GAT (Graph Attention Network)]
Same setup as GraphSAGE, using the implementation provided by PyTorch Geometric (PyG).  
Embeddings were trained on the bipartite co-occurrence graph, and evaluation used the same \(Q(c=1 \mid i,j)\) as above without additional calibration.
\end{description}

Among all baselines, only SGNS required calibration, due to its strictly non-negative inner products resulting from SPPMI truncation.  
Other methods either produced well-distributed inner products or were harmed by additional fitting.

\subsection*{B-6. SVO co-occurrence data}

We used the SVO dataset provided by the NELL project (http://rtw.ml.cmu.edu/resources/svo/), which originally contains 327,255 non-zero triplets over 34,985 subjects, 12,710 verbs, and 46,812 objects.

Due to extreme sparsity (\(\sim 10^{-8}\) density), we first applied Markov diffusion over the entire vocabulary to obtain global co-occurrence signals.  
We then selected 200 frequent and representative words for each of S, V, and O in a step-wise manner (S \(\rightarrow\) V \(\rightarrow\) O), recalculating frequencies at each step.

The resulting \(200^3\) tensor has 509,107 non-zero entries (6.36\% density), which was used for training and evaluation.

We initialized all embeddings using PCA, following the same setup as in previous experiments.

\end{document}